\documentclass{article} % For LaTeX2e
\usepackage{arxiv}
\renewcommand{\headeright}{}
\renewcommand{\undertitle}{}
\usepackage{natbib}
\usepackage{microtype}

\usepackage{amsmath,amsfonts,bm}

\def\eqref#1{equation~\ref{#1}}
\def\1{\bm{1}}

\DeclareMathAlphabet{\mathsfit}{\encodingdefault}{\sfdefault}{m}{sl}
\SetMathAlphabet{\mathsfit}{bold}{\encodingdefault}{\sfdefault}{bx}{n}

\usepackage{graphicx}
\usepackage{xcolor}
\usepackage{algorithm}
\usepackage{algorithmic}
\usepackage{hyperref}
\usepackage{url}
\usepackage{booktabs}
\usepackage{multirow} 
\usepackage{makecell}
\usepackage{amssymb}
\usepackage[framemethod=default]{mdframed}
\usepackage[strict]{changepage}
\usepackage{enumitem}

\definecolor{darkblue}{rgb}{0, 0, 0.5}
\hypersetup{colorlinks=true, citecolor=darkblue, linkcolor=darkblue, urlcolor=darkblue}

\newmdenv[
  linewidth=0pt,
  linecolor=black,
  innerleftmargin=5pt,
  innerrightmargin=5pt,
  skipabove=5pt,
  skipbelow=5pt
]{promptbox}

\newmdenv[
  linewidth=1pt,
  linecolor=black,
  topline=true,
  bottomline=true,
  leftline=true,
  rightline=true,
  innerleftmargin=10pt,
  innerrightmargin=10pt,
  innertopmargin=10pt,
  innerbottommargin=10pt,
  skipabove=1pt,
  skipbelow=1pt
]{examplebox}

\title{SimulRAG: Simulator-based RAG for Grounding LLMs in Long-form Scientific QA}

\author{
\begin{tabular}{@{}c@{}c@{}}
\makebox[0.48\textwidth][c]{%
\begin{tabular}[t]{c}
\rule{0pt}{24pt}Haozhou Xu$^{*}$\\
{\normalfont\small University of California San Diego}\\
{\normalfont\small La Jolla, CA}\\
{\normalfont\small\texttt{hax040@ucsd.edu}}
\end{tabular}}
&
\makebox[0.48\textwidth][c]{%
\begin{tabular}[t]{c}
\rule{0pt}{24pt}Dongxia Wu$^{*}$\\
{\normalfont\small Mohamed bin Zayed University of Artificial Intelligence}\\
{\normalfont\small Abu Dhabi, United Arab Emirates}\\
{\normalfont\small\texttt{dongxia.wu@mbzuai.ac.ae}}
\end{tabular}}
\\
\makebox[0.48\textwidth][c]{%
\begin{tabular}[t]{c}
\rule{0pt}{24pt}Matteo Chinazzi\\
{\normalfont\small Network Science Institute, Northeastern University}\\
{\normalfont\small Portland, ME}\\
{\normalfont\small\texttt{m.chinazzi@northeastern.edu}}
\end{tabular}}
&
\makebox[0.48\textwidth][c]{%
\begin{tabular}[t]{c}
\rule{0pt}{24pt}Ruijia Niu\\
{\normalfont\small University of California San Diego}\\
{\normalfont\small La Jolla, CA}\\
{\normalfont\small\texttt{rniu@ucsd.edu}}
\end{tabular}}
\\
\makebox[0.48\textwidth][c]{%
\begin{tabular}[t]{c}
\rule{0pt}{24pt}Rose Yu\\
{\normalfont\small University of California San Diego}\\
{\normalfont\small La Jolla, CA}\\
{\normalfont\small\texttt{roseyu@ucsd.edu}}
\end{tabular}}
&
\makebox[0.48\textwidth][c]{%
\begin{tabular}[t]{c}
\rule{0pt}{24pt}Yi-An Ma\\
{\normalfont\small University of California San Diego}\\
{\normalfont\small La Jolla, CA}\\
{\normalfont\small\texttt{yianma@ucsd.edu}}
\end{tabular}}
\end{tabular}
}

\date{}

\newcommand{\blfootnote}[1]{\begingroup%
\renewcommand\thefootnote{}\footnotetext{#1}%
\endgroup}

\begin{document}

\maketitle

\blfootnote{$^*$ Equal contribution}

\begin{abstract}
Large Language Models (LLMs) show promise in generating long-form scientific explanations that synthesize evidence and connect multiple factors. However, in long-form scientific question answering, LLMs often hallucinate, producing unsupported or inconsistent claims. Retrieval-Augmented Generation (RAG) improves trustworthiness by grounding generation in external sources; scientific simulators are valuable because they can validate quantitative hypotheses and capture evolving dynamics. Yet simulation-based RAG is non-trivial due to two challenges: how to retrieve from scientific simulators, and how to efficiently verify and update long-form answers. To overcome these challenges, we propose SimulRAG\footnote{Project page: \url{https://github.com/HZEmpire/SimulRAG}}, a simulator-based RAG framework with a generalized retrieval interface that translates between text and simulator parameters/outputs. SimulRAG further introduces claim-level generation with uncertainty estimation and simulator boundary assessment (UE+SBA) to selectively verify and update claims. Unlike tool-first or holistic answer revision, it first elicits diverse answers without retrieval and then grounds uncertain, simulator-verifiable atomic claims with simulator evidence. We also release a long-form scientific QA benchmark spanning climate science, epidemiology, and urban planning, with ground truth verified by simulations and human annotators. Experiments show SimulRAG improves informativeness by 30.4\% and factuality by 16.3\% over the strongest adapted RAG baselines, while UE+SBA enhances claim-level efficiency and quality.
\end{abstract}

\section{Introduction}

The lofty goal of developing AI scientists
has driven extensive LLM research across scientific tasks, ranging
from exam-style question answering (QA)~\citep{lu2022learn, zhang2024sciglm}
to hypothesis proposal~\citep{wang2024scimon, yang2024moose} and
experiment design~\citep{chen2024scienceagentbench, mialon2023gaia}.
Among these, long-form QA is important because answers must blend multiple scientific
claims and reason through complex phenomena from different perspectives~\citep{rein2024gpqa,lee2023qasa}.
For example, predicting disease spread dynamics in epidemiology
requires jointly analyzing transmissibility, incubation, clinical severity, seasonal variations,
population demographics, and contact mixing patterns~\citep{chang2020modelling,cramer2022evaluation}.
However, comprehensive studies on LLMs for long-form scientific
QA remain limited, and existing work on
general free-form QA still manifests persistent hallucination
issues~\citep{farquhar2024detecting}.

Recent works have shown that grounding LLMs with external knowledge sources
can help mitigate hallucination and improve answer factuality~\citep{schick2023toolformer,patil2024gorilla}. In the
scientific domain, it is natural to consider scientific simulators or corresponding
emulators as tools to solve scientific problems~\citep{ren2025towards,ma2024llm}. Compared with static textual knowledge bases
such as literature reviews, querying simulators can capture evolving dynamics and provide
more detailed, timely information about specific scientific phenomena~\citep{ren2025towards}.
Existing works have focused on fine-tuning LLMs to use scientific tools for predefined tasks~\citep{lyu2024adapting,thulke2024climategpt,zhang2024sciglm},
which requires expensive computational resources, predefined question templates,
and precollected datasets, limiting generalizability across scientific domains.

\begin{figure}[t]
\centering
\includegraphics[width=\textwidth]{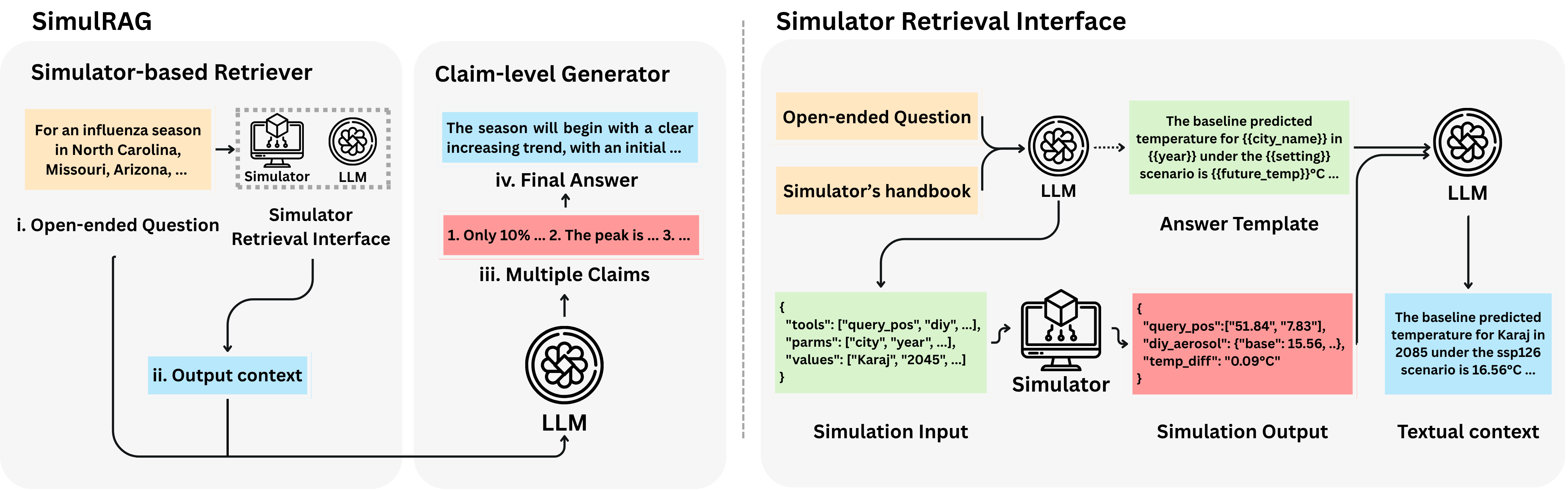}
\caption{Left: Overall SimulRAG structure, including
the simulator-based retriever and claim-level generator.
Right: Simulator retrieval interface:
(1) prompting LLM with question and handbook to extract
simulator parameter settings; (2) executing simulator with parameters
to obtain simulation outputs; (3) converting outputs to
textual context via predefined or LLM-generated templates.}
\label{fig:simulrag_framework}
\end{figure}

Retrieval-Augmented Generation (RAG) approaches have shown promise for enhancing
LLMs by incorporating external knowledge sources to
improve answer factuality and informativeness~\citep{lewis2020retrieval, fan2024survey, gao2023retrieval} without
additional fine-tuning, making them strong candidates for grounding LLMs as trustworthy AI scientists.
RAG consists of two main components: a \textit{retriever} that searches relevant documents
and a \textit{generator} that produces answers based on the retrieved content.
Unfortunately, two fundamental challenges hinder applying existing
RAG approaches to simulation-based retrieval for long-form QA in scientific domains.
First, the discrepancy between textual space and the numerical space where simulation parameters and outputs reside
creates difficulties for querying scientific simulators. Second, existing RAG generators cannot effectively
update long-form answers with new context because they lack fine-grained control.

To overcome these challenges, we introduce the simulator-based RAG framework (SimulRAG)
for long-form scientific QA. The overall structure is
illustrated in Figure~\ref{fig:simulrag_framework} left. It provides
a generalized simulator retrieval interface to transform between textual and numerical
modalities, enabling seamless integration of scientific simulators into RAG systems.
We further introduce
a granular generation method that decomposes long-form answers into atomic
claims and verifies or updates each claim. To improve generation efficiency,
we utilize uncertainty estimation scores
and simulator boundary assessment (UE+SBA) to verify and update claims only when necessary.
To systematically evaluate long-form scientific QA, we additionally construct a benchmark
using simulators as retrieval tools in climate modeling, epidemiology and urban planning, with
ground truth verified by both simulations and human annotators.
This claim-first design preserves answer coverage while focusing costly
claim verification on uncertain claims within the simulator's verification boundary.
Separating answer elicitation from evidence-based correction also allows verification
effort to be allocated at the claim level rather than rewriting an entire response after retrieval.
The extensive experimental results demonstrate that SimulRAG outperforms traditional
RAG baselines adapted with simulator retrieval interface in factuality and informativeness, while UE+SBA
improves claim-level generation efficiency and quality.

Our contributions are summarized as follows:
\begin{itemize}
\item We introduce SimulRAG, a simulator-based RAG framework
for long-form scientific QA.
\item We propose a generalized simulator retrieval interface to transform
between textual and numerical modalities.
\item We present a claim-level generation method to improve long-form answer quality.
\item We utilize uncertainty estimation scores
and simulator boundary assessment (UE+SBA) to efficiently verify and update claims.
\item We construct a long-form scientific QA benchmark for climate science, epidemiology and urban planning.
\item We conduct extensive experiments to verify SimulRAG
framework and UE+SBA method effectiveness. Results show
SimulRAG outperforms RAG baselines by 30.4\% in
informativeness and 16.3\% in factuality.
\end{itemize}

\begin{figure}[H]
\centering
\includegraphics[width=\textwidth]{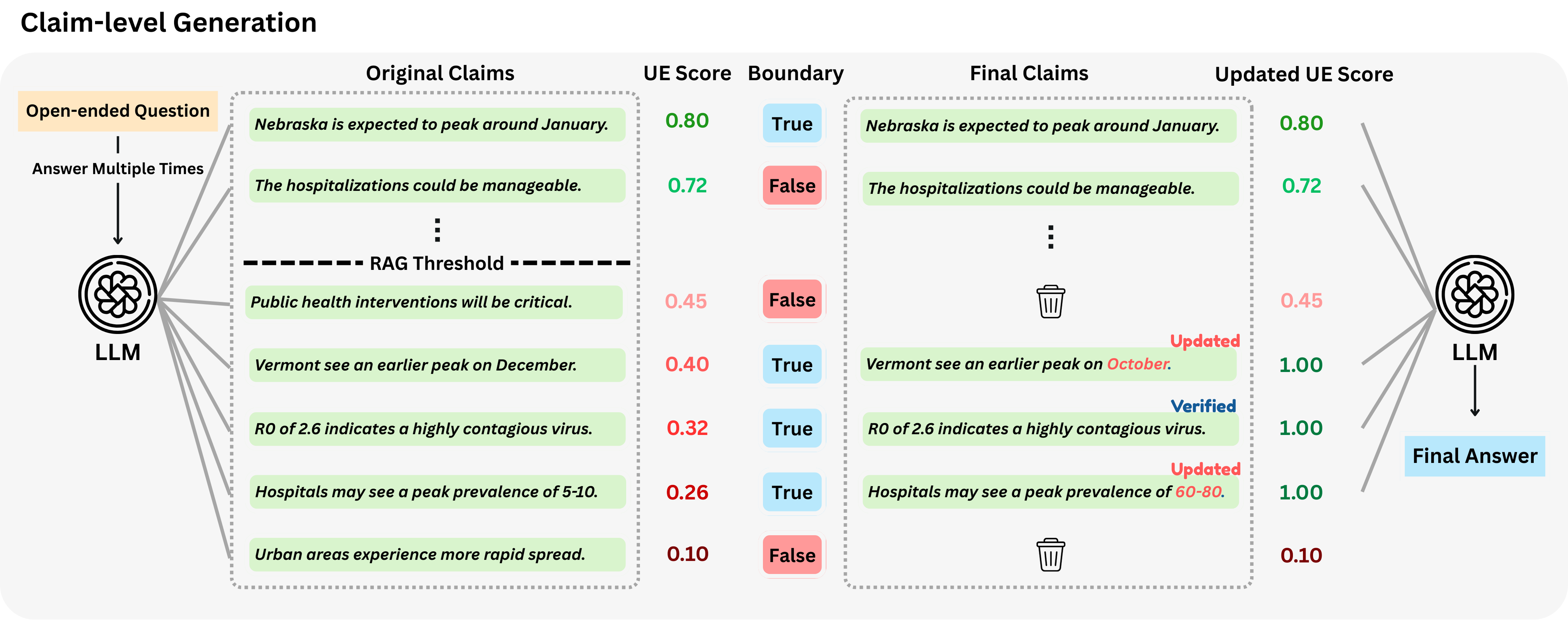}
\caption{Claim-level generation process: (1) decompose long-form answers
into atomic claims; (2) apply uncertainty estimation and
simulator boundary assessment to select claims for verification;
(3) update selected claims using simulation context;
(4) integrate verified claims into a coherent answer.}
\label{fig:claim_level_generation}
\end{figure}

\section{Related work}

\subsection{Scientific question answering}
Scientific question answering encompasses multiple formats and domains.
ScienceQA~\citep{lu2022learn} and Microvqa~\citep{burgess2025microvqa}
focus on multimodal multiple-choice reasoning rather than
free-form QA.
Climate Crisis QA~\citep{zhu2023climate} and SciQAG-24D~\citep{wan2024sciqag}
explore synthetic data generation using LLMs,
but may suffer from hallucinations and limited
scientific validity.
CLIMAQA~\citep{manivannan2024climaqa} provides automated evaluation frameworks
for climate science QA.
SciQA~\citep{auer2023sciqa} benchmarks scientific QA using
hand-crafted queries on the Open Research
Knowledge Graph~\citep{jaradeh2019open}.
Both support free-form QA but address short-form questions whose
answers are typically simple and contain a single claim.
Our questions instead require multiple claims forming complete answers.
We provide the benchmark for long-form scientific QA across climate
modeling, epidemiological modeling and urban planning. The ground truth answers
are verified by both scientific simulators and human annotators.

\subsection{Retrieval-augmented generation (RAG) for LLMs}
RAG systems consist of retrieval and generation components~\citep{lewis2020retrieval, fan2024survey, gao2023retrieval}.
Retrieval is commonly implemented with dense methods that map
queries and knowledge into vector spaces~\citep{khandelwal2019generalization,karpukhin2020dense,lewis2020pre},
or sparse methods based on word matching~\citep{robertson2009probabilistic,sparck1972statistical}.
Generation can incorporate retrieval at the input layer~\citep{ram2023context,izacard2020leveraging},
output layer~\citep{khandelwal2019generalization,yogatama2021adaptive,yu2023improving},
or intermediate layer~\citep{borgeaud2022improving,wu2022memorizing}, though intermediate approaches require
model access unavailable through many LLM APIs~\citep{ma2023query}.
Existing RAG methods rely on search-based retrieval mechanisms
that are unsuitable for scientific simulators.
Recent work explores scientific tool integration~\citep{lyu2024adapting,wang2024scimon,yang2024moose,majumder2024discoverybench},
but mostly targets predefined tasks rather than free-form
scientific QA with open-ended questions.
Our work proposes a general RAG framework
using simulators as retrieval tools.

\subsection{Claim-level uncertainty estimation}
Long-form answer uncertainty estimation decomposes answers into
multiple claims for granular assessment.
Several methods~\citep{duan2023shifting,band2024linguistic} obtain claim uncertainty
scores from long-form outputs but require
white-box model access, limiting applicability
to API-based LLMs.
SelfCheckGPT~\citep{manakul2023selfcheckgpt} extends self-consistency~\citep{wang2022self}
to sentence-level uncertainty for black-box LLMs.
Mohri et al.~\citep{mohri2024language} perform claim-level uncertainty
estimation using conformal prediction, and Jiang et al.~\citep{jiang2024graph}
improve granular estimation through entailment graphs capturing
fine-grained semantic information.
We adapt this claim-level uncertainty method
to guide claim-level generation,
improving its efficiency and quality.

\section{Methodology}

\subsection{Overall framework}
Our proposed simulator-based RAG (SimulRAG) framework
retrieves from scientific simulators to ground LLMs for
long-form scientific QA tasks.
Given an open-ended question $q$, a simulator $S$, and its handbook $h$,
SimulRAG first produces initial long-form answers and decomposes them into atomic claims, then retrieves simulator outputs $o$ via a simulator retrieval interface $I$ for subsequent claim verification and updating:
\begin{equation}
o = I(S, q, h).
\end{equation}
The raw outputs $o$ are converted into textual context $d$ via templates. The generator $G$ takes
$q$, $d$, and LLM model $M$ to produce the final claim set $\mathcal{C}=\{c_1, c_2, \ldots, c_k\}$ after claim-level verification and updating:
\begin{equation}
\mathcal{C} = G(q, d, M)
\end{equation}
where claim set $\mathcal{C}$ contains multiple atomic claims that form the final long-form answer $a$.

The overall SimulRAG framework is illustrated in Figure~\ref{fig:simulrag_framework} left.
Our main contributions in this framework are the design of simulator retrieval interface $I$
and the claim-level generator $G$. For the simulator retrieval interface $I$, we design a generalized
approach to transform between textual and numerical modalities. For the claim-level
generator $G$, we utilize uncertainty estimation scores
and simulator boundary assessment (UE+SBA) to efficiently verify and update claims. The
details of $I$ and $G$ are described in the following sections.
The complete algorithm is presented in Algorithm~\ref{alg:simulrag_framework}.

\subsection{Simulator retrieval interface}
\label{subsec:simulator_retrieval}
Our simulator retrieval interface transforms between textual and
numerical modalities. Figure~\ref{fig:simulrag_framework} right illustrates the overall
process. We use question $q$ and simulator $S$'s
handbook as context $h$ to guide the LLM
in understanding simulator functionality and parameter space. The
LLM extracts multiple relevant parameter settings from question
$q$ and context $h$. These parameters are transformed
to JSON format and executed by simulator $S$
to obtain outputs $o$. Simulation outputs are verbalized using
predefined or LLM-generated templates. Adapting a new simulator only
requires a lightweight adapter specifying its functions, parameter schema,
and outputs; the retrieval and claim-verification pipeline remains unchanged.
Prompts, templates, and the detailed retrieval procedure are provided in Appendix~\ref{sec:appendix_prompt_template} and Algorithm~\ref{alg:retrieval_interface}.

\subsection{Claim-level generation}
\label{subsec:claim_level_generation}

Traditional RAG generation methods directly produce answers given
questions and retrieved context. This one-step mechanism often
yields suboptimal informativeness and factuality. To improve this,
we generate multiple diverse answers $\mathcal{A}=\{a_1, a_2, \ldots, a_m\}$
through $m$ LLM queries, covering different aspects
of the question. Each answer $a_i$ is
decomposed into atomic claims $\{c_{i1}, c_{i2}, \ldots, c_{in}\}$
following \citep{min2023factscore}, where each claim $c_{ij}$
represents an independently verifiable factual statement.
Claims from different answers merge into a
single deduplicated set $\mathcal{C}=\{c_1, c_2, \ldots, c_k\}$
using the LLM deduplication approach from \citep{jiang2024graph}.

This claim-level decomposition serves two critical purposes: (1)
enabling targeted verification and updates of individual
claims rather than holistic response modification, providing
more precise and flexible long-form answer refinement;
(2) simplifying the verification task by focusing
on atomic factual statements. Each claim represents a concise
scientific assertion about phenomena, relationships, or quantitative
predictions that can be directly validated against
simulation outputs.

However, verifying all $k$ claims requires $O(k)$
verification queries, creating computational bottlenecks. We address
this through uncertainty estimation scores and simulator
boundary assessment (UE+SBA), which selectively verify only
uncertain and verifiable claims. Figure~\ref{fig:claim_level_generation}
illustrates the complete claim-level generation process.

\textbf{Claim-level Uncertainty Estimation}
To assess the uncertainty estimation score of each claim $c_i$, we construct bipartite graphs
between the answer set $\mathcal{A}$ and the claim set $\mathcal{C}$.
Each node represents either an answer or a claim, while edges
capture semantic entailment relationships between them. We estimate claim-level
uncertainty using graph centrality metrics, specifically adopting closeness centrality
which measures how close a claim node is to all other nodes.
\begin{equation}
\text{conf}(c_i) = \frac{|\mathcal{V}| - 1}{\sum_{u \in \mathcal{V}} d(c_i, u)} \cdot \frac{|\mathcal{V}|}{|\mathcal{V}_{c_i}|}
\end{equation}
where $\mathcal{V}$ is the set of all nodes, $d(c_i, u)$ is
the shortest path distance between claim node $c_i$ and node $u$,
and $|\mathcal{V}_{c_i}|$ represents the size of the connected
component containing $c_i$. Higher closeness centrality indicates
a more central claim with stronger support from multiple answers,
thus higher confidence. Our approach differs from~\citep{jiang2024graph} in 
purpose: while they detect uncertain claims for abstention,
we identify uncertain claims for verification and modification
through simulation output context to improve answer factuality.
We evaluate additional graph centrality metrics as uncertainty estimators,
with results presented in Section~\ref{sec:ue_sba_results}.

\textbf{Simulator Boundary Assessment}
Another important factor to consider is whether claim $c_i$ can be verified by simulator $S$.
To this end, we introduce a boundary compatibility function 
\begin{equation}
\text{bound}(c_i, h) \rightarrow \{0, 1\}
\end{equation}

This function returns a binary value indicating whether
claim $c_i$ parameters and conditions fall within
simulator $S$ operational boundaries. We employ GPT-4o
as an LLM judge, providing simulator handbook $h$
and claim $c_i$ for evaluation. The judge
determines compatibility between claim parameters and simulator
capabilities, returning 1 for compatible claims and
0 for incompatible ones. This assessment filters
incompatible claims before verification, reducing unnecessary queries
and improving efficiency.

\textbf{UE+SBA Selection and Verification and Updating.}
We combine uncertainty estimation and boundary assessment
for selective claim verification. A claim $c_i$ undergoes
simulator verification when meeting two criteria: (1)
uncertainty: $\text{conf}(c_i) < \tau$ where $\tau$
represents a predefined confidence threshold; (2) boundary
compatibility: $\text{bound}(c_i, h) = 1$. This dual-criterion
approach optimally allocates computational resources to uncertain
yet verifiable claims.

Selected claims undergo verification using simulation context $d$
through three scenarios: (1) alignment: claims consistent
with $d$ retain original content with $\text{conf}(c_i) = 1$;
(2) contradiction: claims conflicting with $d$ are
updated based on simulation evidence with $\text{conf}(c_i) = 1$;
(3) indeterminate: unverifiable claims preserve original
content and confidence scores. Finally, we apply
confidence threshold $\kappa$ to filter low-confidence
claims. The high-confidence set $\{c_i \in \mathcal{C} | \text{conf}(c_i) \geq \kappa\}$
integrates into the final answer $a'$.

\begin{figure}[t]
\centering
\includegraphics[width=\textwidth]{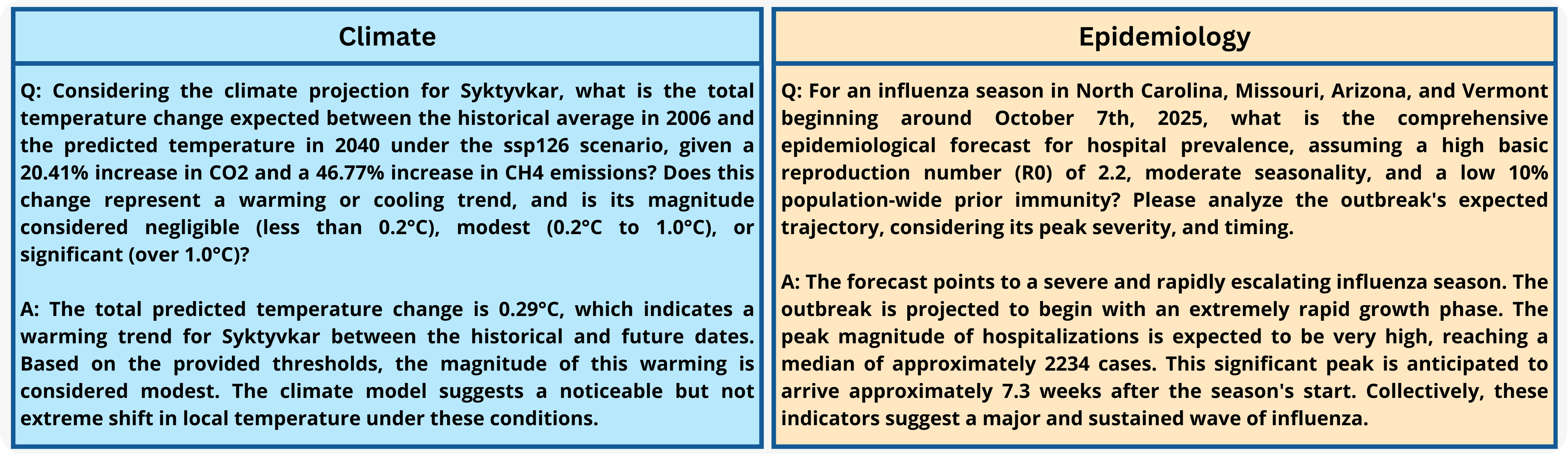}
\caption{Example questions and answers from our benchmark
dataset for climate and epidemiology.}
\label{fig:dataset_overview}
\end{figure}

\subsection{Scientific benchmark generation}
\label{sec:dataset_generation}
We construct a benchmark dataset for long-form
scientific QA using simulators as retrieval tools,
covering climate modeling, epidemiology, and urban planning domains.
For climate modeling, we utilize the climate
emulator~\citep{niu2024multi} trained on Coupled Model Intercomparison
Project Phase 6~\citep{eyring2016overview} simulations, comprising
general circulation and Earth system models
representing the scientific standard for climate projection.
This emulator enables efficient exploration of climate
scenarios that would be prohibitively expensive with full CMIP6 models.
Focusing on four greenhouse and aerosol gases
($CO_2, CH_4, BC, SO_2$) as inputs and 2-meter
surface temperature as output, it captures
anthropogenic warming drivers while targeting a policy-relevant
climate variable. This enables rapid assessment
of emission pathways and climate risk.
% This emulator is able to accurately reproduce the mechanistic disease dynamics of the Global Epidemic and Mobility model (GLEAM) 
% \citep{balcan2010modeling, chinazzi2024multiscale}, a stochastic, age-stratified, metapopulation model that combines high 
% resolution population data, age-stratified social mixing dynamics, human mobility, and disease transmission into 
% a unified computational framework. Our benchmark settings consider a Susceptible-Latent-Infectious-Removed-like 
% compartmental model (details are provided in \citep{zahedi2024gleam}) that is used to simulate the evolution 
% of seasonal influenza outbreaks in the U.S. and that was previously validated in influenza forecasting 
% efforts \citep{mathis2024evaluation}. In this benchmark, we generate the raw data by varying the following parameters 
% in GLEAM-AI: the basic reproduction number (R0), the strength of the seasonality, the level of initial 
% residual immunity in the population, and the presumed starting date of the outbreak.
For epidemiology, we employ GLEAM-AI, a stochastic
emulator that reproduces influenza transmission patterns in the
United States~\citep{zahedi2024gleam, wu2023deep}. It
accurately replicates the mechanistic disease dynamics of 
the Global Epidemic and Mobility model
(GLEAM)~\citep{balcan2010modeling, chinazzi2024multiscale}, a stochastic,
age-stratified, metapopulation model integrating population
data, age-stratified social mixing, human mobility, and disease transmission.
% Our benchmark considers a
% Susceptible-Latent-Infectious-Removed-like compartmental model~\citep{zahedi2024gleam}
% simulating seasonal influenza outbreaks in the U.S., previously validated
% in influenza forecasting efforts~\citep{mathis2024evaluation}.
We vary the following parameters
in GLEAM-AI: basic reproduction number (R0),
the strength of the seasonality, the level of initial 
residual immunity in the population, and the presumed starting date of the outbreak. For urban planning, we utilize the SUMO simulator~\citep{lopez2018microscopic} to model traffic dynamics under interventions including signal prioritization, modal shifts to cycling, and arterial closures, and assess their impacts on travel time efficiency, vehicle idling, and $CO_2$ emissions.

Using these emulators, we design a three-stage
benchmark generation pipeline. First, the LLM receives
simulator handbooks and identifies core functionalities, generating
textual templates describing input-output relationships with placeholders for input
parameters and simulation results stored in
JSON format. Second, we programmatically sample
input parameters from scientifically plausible ranges,
execute simulators, and populate template placeholders
with parameter-output pairs. Finally, we prompt
the LLM to formulate open-ended questions
requiring quantitative reasoning and qualitative interpretation,
directly referencing numerical data. The
LLM then generates ground-truth answers derived solely
from simulation evidence, ensuring factual consistency
without relying on external knowledge.
The sampled simulator parameters and outputs used to construct reference
answers are privileged benchmark-generation information and are not exposed
at test time. At test time, a method receives only the question and simulator
interface.
Figure~\ref{fig:dataset_overview} shows example questions and answers.

\section{Experiments}
\subsection{Experimental setup}
We evaluate our proposed SimulRAG framework and UE+SBA method
across three scientific domains: climate, epidemiology, and urban planning.
We employ GPT-4o and Claude-3.5 Sonnet as backbone LLMs.\footnote{Claude-Haiku-4.5 replaces Claude-3.5 in some Urban experiments because the latter was no longer available.}
Our objective is to demonstrate that SimulRAG improves answer
informativeness and factuality compared to traditional RAG
baselines, while UE+SBA enhances efficiency and quality for claim-level generation.

\begin{figure*}[t]
\centering
\includegraphics[width=\textwidth]{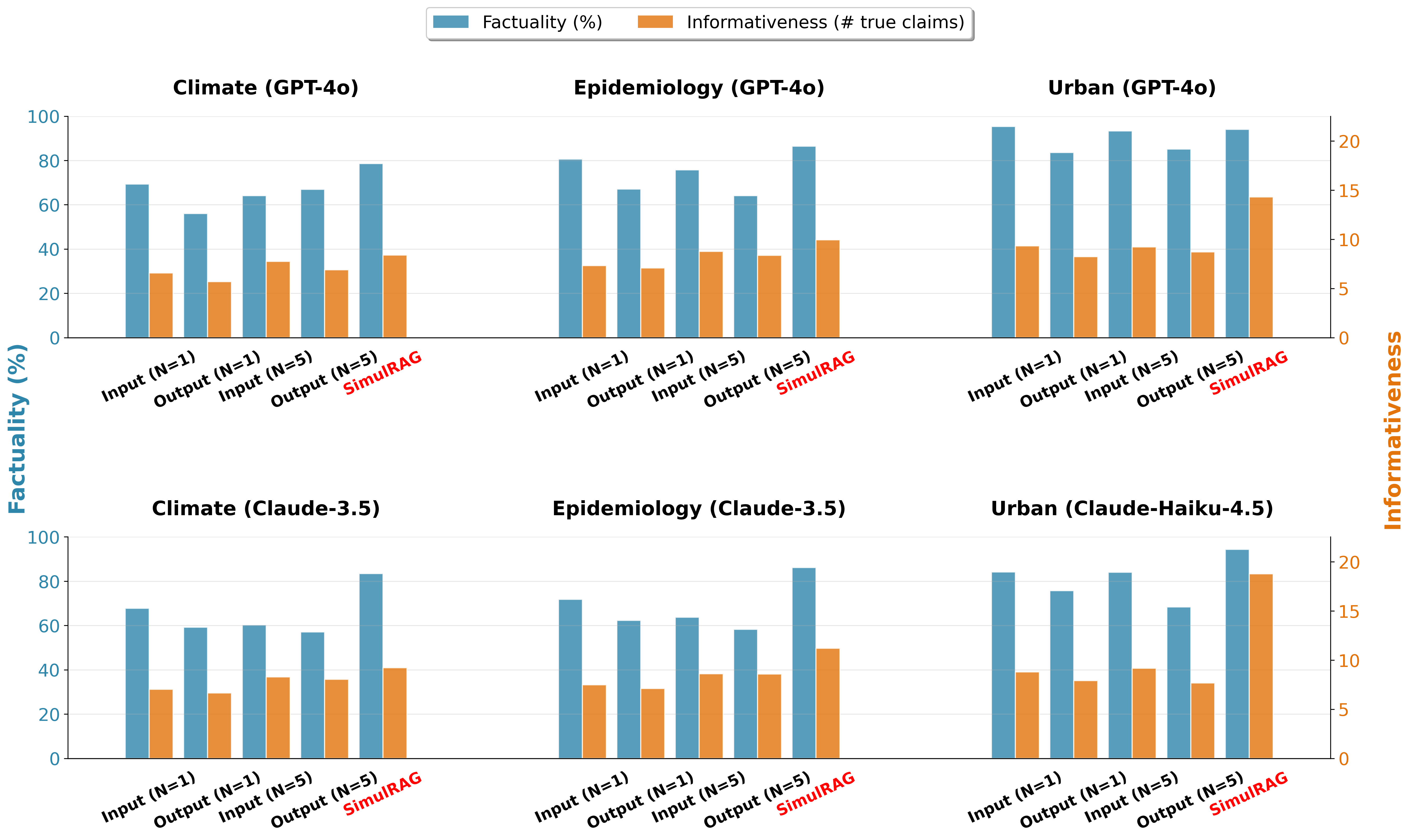}
\caption{Factuality and informativeness comparison across RAG methods. Input- and output-layer baselines sample $N=1$ or $N=5$ answers, while SimulRAG uses $m=5$ initial answers. Across domains and models, SimulRAG substantially improves informativeness while maintaining competitive factuality, yielding the strongest overall trade-off.}
\label{fig:simulrag_results}
\end{figure*}

\begin{figure*}[t]
\includegraphics[width=\textwidth]{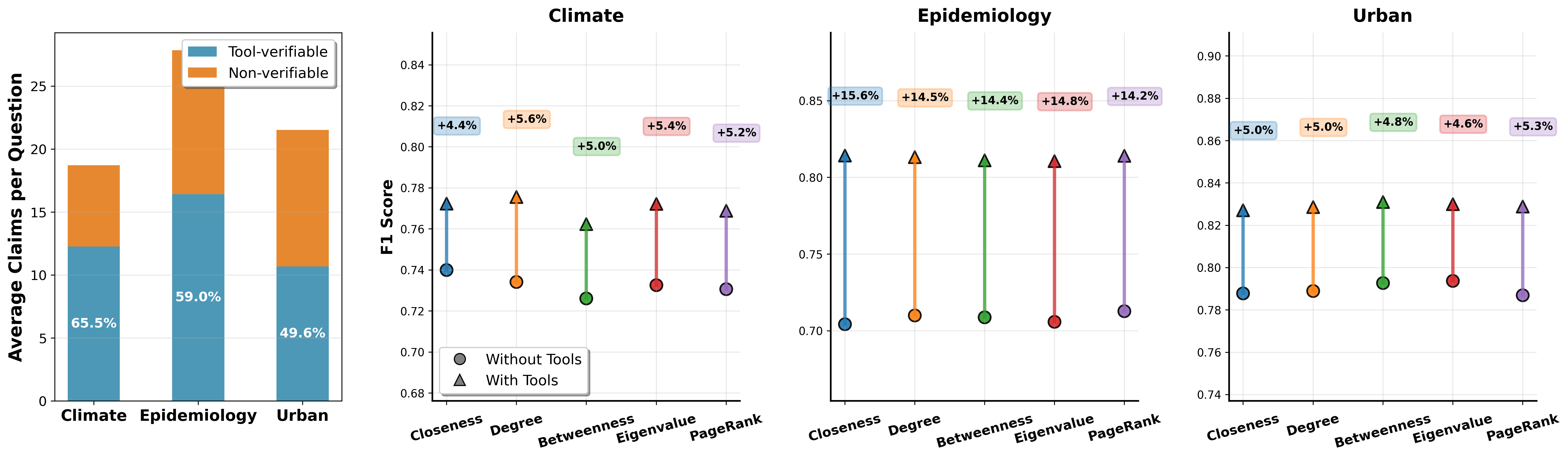}
\caption{Left: Proportion of SBA-selected claims verifiable by simulator.
Right: Performance comparison between Uncertainty and UE+SBA across
five uncertainty estimation scores on climate science, epidemiology, and urban planning benchmarks.}
\label{fig:sba_effectiveness}
\end{figure*}

\begin{table*}[t!]
\centering
\small
\setlength{\tabcolsep}{2.8pt}
\renewcommand{\arraystretch}{1.15}
\resizebox{\textwidth}{!}{%
\begin{tabular}{lll||ccc|ccc|ccc}
\toprule
\multirow{2}{*}{\textbf{Benchmark}} & \multirow{2}{*}{\textbf{Model}} & \multirow{2}{*}{\textbf{Method}} & \multicolumn{3}{c|}{\textbf{F1 Score}} & \multicolumn{3}{c|}{\textbf{AUPR}} & \multicolumn{3}{c}{\textbf{AUROC}} \\
& & & 15\% & 25\% & 45\% & 15\% & 25\% & 45\% & 15\% & 25\% & 45\% \\
\midrule
\multirow{8}{*}{Climate}
& \multirow{4}{*}{GPT-4o}
  & Random & 0.6685 & 0.6794 & 0.7039 & 0.7144 & 0.7383 & 0.7594 & 0.6013 & 0.6267 & 0.6618 \\
& & Verbalized & 0.6519 & 0.6470 & 0.7393 & 0.6745 & 0.7071 & 0.7461 & 0.5522 & 0.6006 & 0.6659 \\
& & Uncertainty & 0.6894 & 0.7113 & 0.7440 & 0.7391 & 0.7522 & 0.7714 & 0.6405 & 0.6645 & 0.7040 \\
& & UE+SBA & \bf{0.7020} & \bf{0.7310} & \bf{0.7725} & \bf{0.7460} & \bf{0.7629} & \bf{0.7871} & \bf{0.6531} & \bf{0.6854} & \bf{0.7253} \\
\cmidrule{2-12}
& \multirow{4}{*}{Claude 3.5}
  & Random & 0.7323 & 0.7419 & 0.7511 & 0.7876 & 0.7932 & 0.8048 & 0.6672 & 0.6793 & 0.7031 \\
& & Verbalized & 0.6766 & 0.7251 & 0.7230 & 0.7532 & 0.7731 & 0.8023 & 0.6172 & 0.6600 & 0.7277 \\
& & Uncertainty & 0.7384 & 0.7509 & 0.7930 & 0.7770 & 0.7817 & 0.8092 & 0.6818 & 0.6975 & 0.7446 \\
& & UE+SBA & \bf{0.7511} & \bf{0.7768} & \bf{0.8242} & \bf{0.7879} & \bf{0.8039} & \bf{0.8297} & \bf{0.7002} & \bf{0.7290} & \bf{0.7709} \\
\midrule
\multirow{8}{*}{Epidemiology}
& \multirow{4}{*}{GPT-4o}
  & Random & 0.5898 & 0.6157 & 0.6581 & 0.6727 & 0.7053 & 0.7555 & 0.6454 & 0.6753 & 0.7295 \\
& & Verbalized & 0.5654 & 0.6270 & 0.6953 & 0.5972 & 0.6534 & 0.7402 & 0.6295 & 0.6834 & 0.7773 \\
& & Uncertainty & 0.6103 & 0.6440 & 0.7043 & 0.6833 & 0.7243 & 0.7799 & 0.6789 & 0.7325 & 0.8005 \\
& & UE+SBA & \bf{0.6431} & \bf{0.6957} & \bf{0.8155} & \bf{0.7239} & \bf{0.7754} & \bf{0.8424} & \bf{0.7308} & \bf{0.7906} & \bf{0.8634} \\
\cmidrule{2-12}
& \multirow{4}{*}{Claude 3.5}
  & Random & 0.6961 & 0.7042 & 0.7374 & 0.7428 & 0.7687 & 0.8097 & 0.5716 & 0.6045 & 0.6723 \\
& & Verbalized & 0.6804 & 0.7421 & 0.7310 & 0.7221 & 0.7599 & 0.8206 & 0.5358 & 0.5993 & 0.7175 \\
& & Uncertainty & 0.7060 & 0.7319 & 0.7848 & 0.7482 & 0.7798 & 0.8260 & 0.5972 & 0.6489 & 0.7397 \\
& & UE+SBA & \bf{0.7231} & \bf{0.7594} & \bf{0.8207} & \bf{0.7739} & \bf{0.8113} & \bf{0.8590} & \bf{0.6338} & \bf{0.6971} & \bf{0.7916} \\
\midrule
\multirow{8}{*}{Urban}
& \multirow{4}{*}{GPT-4o}
  & Random & 0.7919 & 0.7980 & 0.8105 & 0.8328 & 0.8359 & 0.8548 & 0.6000 & 0.6029 & 0.6280 \\
& & Verbalized & 0.7777 & 0.7859 & 0.8022 & 0.7880 & 0.8021 & 0.8265 & 0.5157 & 0.5304 & 0.5541 \\
& & Uncertainty & 0.7949 & 0.7993 & 0.8151 & 0.8292 & 0.8338 & 0.8529 & 0.5982 & 0.6005 & 0.6223 \\
& & UE+SBA & \bf{0.8078} & \bf{0.8224} & \bf{0.8290} & \bf{0.8444} & \bf{0.8569} & \bf{0.8715} & \bf{0.6346} & \bf{0.6537} & \bf{0.6774} \\
\cmidrule{2-12}
& \multirow{4}{*}{Claude 3.5}
  & Random & 0.6801 & 0.7030 & 0.7525 & 0.7575 & 0.7919 & 0.8467 & 0.5892 & 0.6146 & 0.6675 \\
& & Verbalized & 0.6863 & 0.7105 & 0.7545 & 0.7365 & 0.7733 & 0.8329 & 0.5817 & 0.6109 & 0.6657 \\
& & Uncertainty & 0.6839 & 0.7082 & 0.7538 & 0.7604 & 0.7921 & 0.8445 & 0.5914 & 0.6178 & 0.6721 \\
& & UE+SBA & \bf{0.7173} & \bf{0.7613} & \bf{0.8222} & \bf{0.7987} & \bf{0.8469} & \bf{0.9019} & \bf{0.6475} & \bf{0.7091} & \bf{0.7833} \\
\bottomrule
\end{tabular}%
}
\renewcommand{\arraystretch}{1}
\caption{Performance comparison of claim-level generation methods across three benchmarks and two LLM models. Results show F1 Score, AUPR, and AUROC for 3 different verification budgets (15\%, 25\%, 45\%). The 15\% budget means 15\% of claims will be verified and possibly updated. Best performing methods are bolded.}
\label{tab:ue_sba_results_final}
\end{table*}

\textbf{Datasets.} We construct three benchmark datasets
 described in Section~\ref{sec:dataset_generation}.
The climate modeling dataset contains 1000 free-form questions
covering various climate phenomena.
The epidemiological modeling dataset comprises 1000 free-form questions
concerning disease spread dynamics and plausible future scenarios. The urban planning modeling dataset contains 200 questions
on transportation planning problems.
Detailed dataset examples and statistics are provided in Appendix~\ref{sec:appendix_dataset_statistics}.

\textbf{Claim-level evaluation.} We assess generated answer quality
at the atomic claim level. Each answer undergoes
decomposition into constituent atomic claims via structured prompting.
Claims are then evaluated through
a two-stage verification process. First, an LLM
judge evaluates simulator-related claims using ground truth references.
Subsequently, PhD students in the relevant fields
manually assess remaining claims for
correctness and relevance to the posed question.
Although the references are simulator-derived, model outputs are not
restricted to reference claims and may contain incorrect, irrelevant, or
out-of-scope content. A generated claim is labeled true only when it is both
correct and relevant to the question. We audited LLM-judge reliability with two PhD-level domain experts on 30 questions per domain, achieving 95.8\% agreement across domains.

\textbf{Baselines.} We evaluate SimulRAG against established RAG baselines:
(1) \textbf{Input-layer integration}~\citep{ram2023context} directly concatenates
queries with retrieved context as LLM input;
(2) \textbf{Output-layer integration}~\citep{yu2023improving} refines
generated answers using retrieval context post-processing.
These baselines require adaptation for scientific simulators.
We adapt them through our simulator retrieval interface
to extract textual context from simulation outputs, enabling fair comparison.
To assess a tool-first agentic alternative, we additionally adapt
ReAct~\citep{yao2023react} to invoke the same simulator interface; the dedicated
comparison is reported in Appendix~\ref{subsec:appendix_react}.

For claim-level generation, we evaluate UE+SBA against baselines:
(1) \textbf{Random}: Randomly selects claims for simulator verification.
(2) \textbf{Verbalized}: Uses LLM verbalized uncertainty estimates~\citep{lin2022teaching,tian2023just}.
(3) \textbf{Uncertainty}: Selects claims using confidence scores~\citep{jiang2024graph}.
All methods use identical claim decomposition and
answer regeneration procedures from Section~\ref{subsec:claim_level_generation}.
This ensures fair comparison.

\textbf{Evaluation metrics.} We assess SimulRAG and baselines
using informativeness and factuality metrics for answers.
Informativeness counts unique true claims within answers.
Factuality measures the proportion of true claims
across all generated claims.

For claim-level methods, the verification budget denotes the fraction of
claims selected for simulator-based updating. F1 captures the precision-recall
balance at each budget, while AUPR and AUROC evaluate ranking quality across
thresholds. Together, these metrics measure whether a method prioritizes claims
that benefit from verification under the same update budget.
Corresponding precision-recall values appear in Appendix~\ref{subsec:appendix_precision_recall}.

\subsection{SimulRAG performance}
% \label{sec:simulrag_results}

We first assess the effectiveness of the SimulRAG framework
for long-form scientific QA. Figure~\ref{fig:simulrag_results}
demonstrates the informativeness and factuality of different RAG methods
across Climate, Epidemiology, and Urban. Across all six domain-model
settings, SimulRAG achieves 30.4\% more unique true claims on average
than the strongest input- or output-layer baseline in each setting.
For factuality, SimulRAG attains a 16.3\% higher proportion of true claims
on average than the strongest baseline.
These joint gains show that selective claim-level grounding expands answer
coverage without introducing unsupported content through holistic conditioning or revision.

\begin{figure*}[t]
\centering
\includegraphics[width=\textwidth]{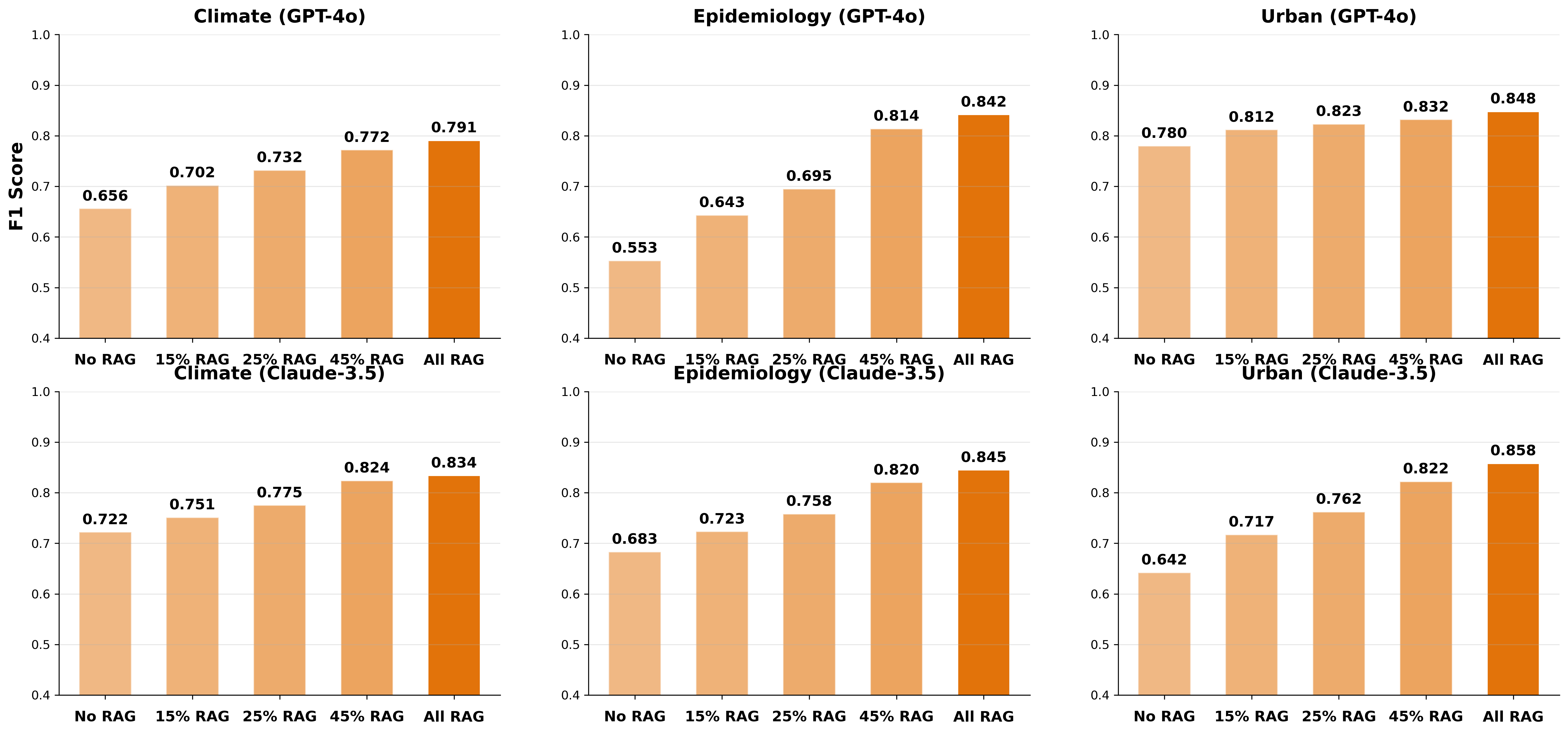}
\caption{F1 score comparison of SimulRAG across RAG
verification budgets (15\%, 25\%, 45\%) versus no-RAG and all-RAG on three benchmarks.}
\label{fig:comparison_without_all_rag}
\end{figure*}

\subsection{UE+SBA performance}
\label{sec:ue_sba_results}
We evaluate the efficiency and quality of uncertainty estimation 
scores and simulator boundary assessment (UE+SBA)
for claim-level generation. Table~\ref{tab:ue_sba_results_final} presents
the performance comparison of different claim-level generation methods
across climate, epidemiology, and urban planning benchmarks, using
GPT-4o and Claude-3.5 models. We report F1 score,
AUPR, and AUROC across 3 verification budgets (15\%, 25\%, 45\%).
UE+SBA consistently outperforms all baseline methods across
all metrics, models, benchmarks, and budgets.
For AUPR, UE+SBA achieves up to 6.2\%
absolute improvements over the best baseline.
For AUROC, UE+SBA attains up to 6.3\%
absolute improvements over the best baseline. 
Meanwhile, the Uncertainty baseline consistently ranks second, demonstrating
the effectiveness of uncertainty estimation for claim
selection, a key component of UE+SBA.
UE+SBA improves F1 consistently across all verification budgets, with its
advantage generally persisting and often increasing as more claims are
selected. This consistency across domains and backbone models indicates that
boundary compatibility provides a complementary selection signal to
uncertainty rather than an improvement tied to a particular operating point.

% For F1 Score, UE+SBA achieves up to 3.1\%
% absolute improvements over the best baseline method.

\subsection{Simulator boundary assessment (SBA) effectiveness}
\label{sec:sba_effectiveness}

To evaluate SBA effectiveness, we visualize the proportion
of SBA-selected claims verifiable by the simulator.
Figure~\ref{fig:sba_effectiveness} left demonstrates that SBA effectively
filters out non-verifiable claims. For climate,
SBA filters out 34.5\% of claims on average.
For epidemiology, SBA filters out 41\% on average.
For urban planning, it filters out 50.4\% on average.
Human evaluation confirms most filtered claims are
indeed non-verifiable by the simulator. Thus, SBA
reduces unnecessary claim-level verification and updates.
Additionally, we compare UE+SBA with Uncertainty 
across five uncertainty estimation scores~\citep{jiang2024graph}.
The five estimators are closeness, degree, betweenness, eigenvector centrality,
and PageRank.
Figure~\ref{fig:sba_effectiveness} right presents the performance
comparison between Uncertainty and UE+SBA across these uncertainty estimation scores.
UE+SBA improves over uncertainty-only selection for each estimator, indicating
that the benefit of boundary filtering is not tied to a particular
graph-centrality definition.

% \begin{figure}[t]
% \centering
% \includegraphics[width=\textwidth]{figs/f1_rag_comparison.png}
% \caption{The performance of different rag ratio of SimulRAG compared to no rag and all rag scenarios}
% \label{fig:framework}
% \end{figure}

\subsection{Ablation studies}

We compare SimulRAG performance across RAG verification
budgets (15\%, 25\%, 45\%) against no-RAG and all-RAG
in Figure~\ref{fig:comparison_without_all_rag} using F1 score.
Results show SimulRAG with 15\% RAG significantly
outperforms no-RAG across the three benchmarks and both models.
Across the six panels, F1 improves over no-RAG by
5.2, 8.5, and 14.1 percentage points at verification budgets of 15\%, 25\%,
and 45\%, respectively. The 45\% setting remains within 3.6\% of all-RAG;
verifying the remaining 55\% of claims adds only 2.2 points on average,
demonstrating diminishing returns beyond the selectively prioritized set.

\section{Conclusion}
In this work, we introduce SimulRAG, a simulator-based retrieval-augmented generation (RAG) framework
for long-form scientific question answering. SimulRAG proposes a generalized simulator retrieval interface
to transform between textual and numerical modalities, enabling seamless integration of scientific simulators
into RAG systems. To improve answer generation quality, we present a claim-level generation method
that decomposes long-form answers into atomic claims for fine-grained verification and updates.
To efficiently verify and update claims, we utilize uncertainty estimation scores
and simulator boundary assessment (UE+SBA) to selectively identify valuable claims for verification.
Finally, we construct a long-form scientific QA benchmark covering climate science, epidemiology, and urban planning.
Extensive experiments verify the effectiveness of our proposed SimulRAG framework and UE+SBA method.
One limitation of SimulRAG is assuming all questions
relate to available simulators. Future work could develop
automatic question-simulator relevance detection to avoid failed retrieval
when prompts are unrelated and enable broader applicability
across diverse scientific domains.

\section*{Acknowledgments}
This work was supported in part by the U.S. Army Research Office
under Army-ECASE award W911NF-23-1-0231, the U.S. Department of Energy, Office of Science, IARPA HAYSTAC Program, CDC-RFA-FT-23-0069, DARPA AIE FoundSci, DARPA YFA, and NSF Grants \#2205093, \#2146343, \#2134274, \#2134209, and \#2112665 (TILOS). M.C. acknowledges support from HHS/CDC 5U01IP0001137, and the cooperative agreement CDC-RFA-FT-23-0069 from the CDC's Center for Forecasting and Outbreak Analytics. The findings and conclusions in this study are those of the authors and do not necessarily represent the official position of the funding agencies, the Centers for Disease Control and Prevention, or the U.S. Department of Health and Human Services.

\bibliography{iclr2026_conference}

@inproceedings{wu2023deep,
  title={Deep bayesian active learning for accelerating stochastic simulation},
  author={Wu, Dongxia and Niu, Ruijia and Chinazzi, Matteo and Vespignani, Alessandro and Ma, Yi-An and Yu, Rose},
  booktitle={Proceedings of the 29th ACM SIGKDD Conference on Knowledge Discovery and Data Mining},
  pages={2559--2569},
  year={2023}
}

@inproceedings{zahedi2024gleam,
  title={GLEAM-AI: Neural Surrogate for Accelerated Epidemic Analytics and Forecasting},
  author={Zahedi, Mohammadmehdi and Wu, Dongxia and Davis, Jessica T and Ma, Yian and Vespignani, Alessandro and Yu, Rose and Chinazzi, Matteo},
  year={2024},
  booktitle={NeurIPS 2024 Workshop on Data-driven and Differentiable Simulations, Surrogates, and Solvers}
}

@article{balcan2010modeling,
  title={Modeling the spatial spread of infectious diseases: The GLobal Epidemic and Mobility computational model},
  author={Balcan, Duygu and Gon{\c{c}}alves, Bruno and Hu, Hao and Ramasco, Jos{\'e} J and Colizza, Vittoria and Vespignani, Alessandro},
  journal={Journal of computational science},
  volume={1},
  number={3},
  pages={132--145},
  year={2010},
  publisher={Elsevier}
}

@article{chinazzi2024multiscale,
  title={A multiscale modeling framework for Scenario Modeling: Characterizing the heterogeneity of the COVID-19 epidemic in the US},
  author={Chinazzi, Matteo and Davis, Jessica T and y Piontti, Ana Pastore and Mu, Kunpeng and Gozzi, Nicol{\`o} and Ajelli, Marco and Perra, Nicola and Vespignani, Alessandro},
  journal={Epidemics},
  volume={47},
  pages={100757},
  year={2024},
  publisher={Elsevier}
}

@inproceedings{fan2024survey,
  title={A survey on rag meeting llms: Towards retrieval-augmented large language models},
  author={Fan, Wenqi and Ding, Yujuan and Ning, Liangbo and Wang, Shijie and Li, Hengyun and Yin, Dawei and Chua, Tat-Seng and Li, Qing},
  booktitle={Proceedings of the 30th ACM SIGKDD conference on knowledge discovery and data mining},
  pages={6491--6501},
  year={2024}
}

@article{eyring2016overview,
  title={Overview of the Coupled Model Intercomparison Project Phase 6 (CMIP6) experimental design and organization},
  author={Eyring, Veronika and Bony, Sandrine and Meehl, Gerald A and Senior, Catherine A and Stevens, Bjorn and Stouffer, Ronald J and Taylor, Karl E},
  journal={Geoscientific Model Development},
  volume={9},
  number={5},
  pages={1937--1958},
  year={2016},
  publisher={Copernicus GmbH}
}

@article{robertson2009probabilistic,
  title={The probabilistic relevance framework: BM25 and beyond},
  author={Robertson, Stephen and Zaragoza, Hugo and others},
  journal={Foundations and Trends{\textregistered} in Information Retrieval},
  volume={3},
  number={4},
  pages={333--389},
  year={2009},
  publisher={Now Publishers, Inc.}
}

@article{sparck1972statistical,
  title={A statistical interpretation of term specificity and its application in retrieval},
  author={Sparck Jones, Karen},
  journal={Journal of documentation},
  volume={28},
  number={1},
  pages={11--21},
  year={1972},
  publisher={MCB UP Ltd}
}

@inproceedings{karpukhin2020dense,
  title={Dense Passage Retrieval for Open-Domain Question Answering.},
  author={Karpukhin, Vladimir and Oguz, Barlas and Min, Sewon and Lewis, Patrick SH and Wu, Ledell and Edunov, Sergey and Chen, Danqi and Yih, Wen-tau},
  booktitle={EMNLP (1)},
  pages={6769--6781},
  year={2020}
}

@article{khandelwal2019generalization,
  title={Generalization through memorization: Nearest neighbor language models},
  author={Khandelwal, Urvashi and Levy, Omer and Jurafsky, Dan and Zettlemoyer, Luke and Lewis, Mike},
  journal={arXiv preprint arXiv:1911.00172},
  year={2019}
}

@article{lewis2020pre,
  title={Pre-training via paraphrasing},
  author={Lewis, Mike and Ghazvininejad, Marjan and Ghosh, Gargi and Aghajanyan, Armen and Wang, Sida and Zettlemoyer, Luke},
  journal={Advances in Neural Information Processing Systems},
  volume={33},
  pages={18470--18481},
  year={2020}
}

@article{wu2022memorizing,
  title={Memorizing transformers},
  author={Wu, Yuhuai and Rabe, Markus N and Hutchins, DeLesley and Szegedy, Christian},
  journal={arXiv preprint arXiv:2203.08913},
  year={2022}
}

@article{ram2023context,
  title={In-context retrieval-augmented language models},
  author={Ram, Ori and Levine, Yoav and Dalmedigos, Itay and Muhlgay, Dor and Shashua, Amnon and Leyton-Brown, Kevin and Shoham, Yoav},
  journal={Transactions of the Association for Computational Linguistics},
  volume={11},
  pages={1316--1331},
  year={2023},
  publisher={MIT Press One Broadway, 12th Floor, Cambridge, Massachusetts 02142, USA~…}
}

@article{izacard2020leveraging,
  title={Leveraging passage retrieval with generative models for open domain question answering},
  author={Izacard, Gautier and Grave, Edouard},
  journal={arXiv preprint arXiv:2007.01282},
  year={2020}
}

@article{yogatama2021adaptive,
  title={Adaptive semiparametric language models},
  author={Yogatama, Dani and de Masson d’Autume, Cyprien and Kong, Lingpeng},
  journal={Transactions of the Association for Computational Linguistics},
  volume={9},
  pages={362--373},
  year={2021},
  publisher={MIT Press One Rogers Street, Cambridge, MA 02142-1209, USA journals-info~…}
}

@article{yu2023improving,
  title={Improving language models via plug-and-play retrieval feedback},
  author={Yu, Wenhao and Zhang, Zhihan and Liang, Zhenwen and Jiang, Meng and Sabharwal, Ashish},
  journal={arXiv preprint arXiv:2305.14002},
  year={2023}
}

@inproceedings{borgeaud2022improving,
  title={Improving language models by retrieving from trillions of tokens},
  author={Borgeaud, Sebastian and Mensch, Arthur and Hoffmann, Jordan and Cai, Trevor and Rutherford, Eliza and Millican, Katie and Van Den Driessche, George Bm and Lespiau, Jean-Baptiste and Damoc, Bogdan and Clark, Aidan and others},
  booktitle={International conference on machine learning},
  pages={2206--2240},
  year={2022},
  organization={PMLR}
}

@inproceedings{ma2023query,
  title={Query rewriting in retrieval-augmented large language models},
  author={Ma, Xinbei and Gong, Yeyun and He, Pengcheng and Zhao, Hai and Duan, Nan},
  booktitle={Proceedings of the 2023 Conference on Empirical Methods in Natural Language Processing},
  pages={5303--5315},
  year={2023}
}

@article{lyu2024adapting,
  title={Adapting While Learning: Grounding LLMs for Scientific Problems with Intelligent Tool Usage Adaptation},
  author={Lyu, Bohan and Cao, Yadi and Watson-Parris, Duncan and Bergen, Leon and Berg-Kirkpatrick, Taylor and Yu, Rose},
  journal={arXiv preprint arXiv:2411.00412},
  year={2024}
}

@inproceedings{wang2024scimon,
  title={Scimon: Scientific inspiration machines optimized for novelty},
  author={Wang, Qingyun and Downey, Doug and Ji, Heng and Hope, Tom},
  booktitle={Proceedings of the 62nd Annual Meeting of the Association for Computational Linguistics (Volume 1: Long Papers)},
  pages={279--299},
  year={2024}
}

@article{majumder2024discoverybench,
  title={Discoverybench: Towards data-driven discovery with large language models},
  author={Majumder, Bodhisattwa Prasad and Surana, Harshit and Agarwal, Dhruv and Mishra, Bhavana Dalvi and Meena, Abhijeetsingh and Prakhar, Aryan and Vora, Tirth and Khot, Tushar and Sabharwal, Ashish and Clark, Peter},
  journal={arXiv preprint arXiv:2407.01725},
  year={2024}
}

@article{lu2022learn,
  title={Learn to explain: Multimodal reasoning via thought chains for science question answering},
  author={Lu, Pan and Mishra, Swaroop and Xia, Tanglin and Qiu, Liang and Chang, Kai-Wei and Zhu, Song-Chun and Tafjord, Oyvind and Clark, Peter and Kalyan, Ashwin},
  journal={Advances in Neural Information Processing Systems},
  volume={35},
  pages={2507--2521},
  year={2022}
}

@inproceedings{burgess2025microvqa,
  title={Microvqa: A multimodal reasoning benchmark for microscopy-based scientific research},
  author={Burgess, James and Nirschl, Jeffrey J and Bravo-S{\'a}nchez, Laura and Lozano, Alejandro and Gupte, Sanket Rajan and Galaz-Montoya, Jesus G and Zhang, Yuhui and Su, Yuchang and Bhowmik, Disha and Coman, Zachary and others},
  booktitle={Proceedings of the Computer Vision and Pattern Recognition Conference},
  pages={19552--19564},
  year={2025}
}

@article{zhu2023climate,
  title={Climate change from large language models},
  author={Zhu, Hongyin and Tiwari, Prayag},
  journal={arXiv preprint arXiv:2312.11985},
  year={2023}
}

@article{wan2024sciqag,
  title={SciQAG: A framework for auto-generated science question answering dataset with fine-grained evaluation},
  author={Wan, Yuwei and Liu, Yixuan and Ajith, Aswathy and Grazian, Clara and Hoex, Bram and Zhang, Wenjie and Kit, Chunyu and Xie, Tong and Foster, Ian},
  journal={arXiv preprint arXiv:2405.09939},
  year={2024}
}

@article{manivannan2024climaqa,
  title={ClimaQA: An Automated Evaluation Framework for Climate Question Answering Models},
  author={Manivannan, Veeramakali Vignesh and Jafari, Yasaman and Eranky, Srikar and Ho, Spencer and Yu, Rose and Watson-Parris, Duncan and Ma, Yian and Bergen, Leon and Berg-Kirkpatrick, Taylor},
  journal={arXiv preprint arXiv:2410.16701},
  year={2024}
}

@article{auer2023sciqa,
  title={The sciqa scientific question answering benchmark for scholarly knowledge},
  author={Auer, S{\"o}ren and Barone, Dante AC and Bartz, Cassiano and Cortes, Eduardo G and Jaradeh, Mohamad Yaser and Karras, Oliver and Koubarakis, Manolis and Mouromtsev, Dmitry and Pliukhin, Dmitrii and Radyush, Daniil and others},
  journal={Scientific Reports},
  volume={13},
  number={1},
  pages={7240},
  year={2023},
  publisher={Nature Publishing Group UK London}
}

@inproceedings{jaradeh2019open,
  title={Open research knowledge graph: Next generation infrastructure for semantic scholarly knowledge},
  author={Jaradeh, Mohamad Yaser and Oelen, Allard and Farfar, Kheir Eddine and Prinz, Manuel and D'Souza, Jennifer and Kismih{\'o}k, G{\'a}bor and Stocker, Markus and Auer, S{\"o}ren},
  booktitle={Proceedings of the 10th international conference on knowledge capture},
  pages={243--246},
  year={2019}
}

@article{duan2023shifting,
  title={Shifting attention to relevance: Towards the predictive uncertainty quantification of free-form large language models},
  author={Duan, Jinhao and Cheng, Hao and Wang, Shiqi and Zavalny, Alex and Wang, Chenan and Xu, Renjing and Kailkhura, Bhavya and Xu, Kaidi},
  journal={arXiv preprint arXiv:2307.01379},
  year={2023}
}

@article{band2024linguistic,
  title={Linguistic calibration of long-form generations},
  author={Band, Neil and Li, Xuechen and Ma, Tengyu and Hashimoto, Tatsunori},
  journal={arXiv preprint arXiv:2404.00474},
  year={2024}
}

@article{manakul2023selfcheckgpt,
  title={Selfcheckgpt: Zero-resource black-box hallucination detection for generative large language models},
  author={Manakul, Potsawee and Liusie, Adian and Gales, Mark JF},
  journal={arXiv preprint arXiv:2303.08896},
  year={2023}
}

@article{wang2022self,
  title={Self-consistency improves chain of thought reasoning in language models},
  author={Wang, Xuezhi and Wei, Jason and Schuurmans, Dale and Le, Quoc and Chi, Ed and Narang, Sharan and Chowdhery, Aakanksha and Zhou, Denny},
  journal={arXiv preprint arXiv:2203.11171},
  year={2022}
}

@article{mohri2024language,
  title={Language models with conformal factuality guarantees},
  author={Mohri, Christopher and Hashimoto, Tatsunori},
  journal={arXiv preprint arXiv:2402.10978},
  year={2024}
}

@article{jiang2024graph,
  title={Graph-based Uncertainty Metrics for Long-form Language Model Generations},
  author={Jiang, Mingjian and Ruan, Yangjun and Sattigeri, Prasanna and Roukos, Salim and Hashimoto, Tatsunori B},
  journal={Advances in Neural Information Processing Systems},
  volume={37},
  pages={32980--33006},
  year={2024}
}

@inproceedings{rein2024gpqa,
  title={Gpqa: A graduate-level google-proof q\&a benchmark},
  author={Rein, David and Hou, Betty Li and Stickland, Asa Cooper and Petty, Jackson and Pang, Richard Yuanzhe and Dirani, Julien and Michael, Julian and Bowman, Samuel R},
  booktitle={First Conference on Language Modeling},
  year={2024}
}

@inproceedings{lee2023qasa,
  title={Qasa: advanced question answering on scientific articles},
  author={Lee, Yoonjoo and Lee, Kyungjae and Park, Sunghyun and Hwang, Dasol and Kim, Jaehyeon and Lee, Hong-in and Lee, Moontae},
  booktitle={International Conference on Machine Learning},
  pages={19036--19052},
  year={2023},
  organization={PMLR}
}

@article{chang2020modelling,
  title={Modelling transmission and control of the COVID-19 pandemic in Australia},
  author={Chang, Sheryl L and Harding, Nathan and Zachreson, Cameron and Cliff, Oliver M and Prokopenko, Mikhail},
  journal={Nature communications},
  volume={11},
  number={1},
  pages={5710},
  year={2020},
  publisher={Nature Publishing Group UK London}
}

@article{cramer2022evaluation,
  title={Evaluation of individual and ensemble probabilistic forecasts of COVID-19 mortality in the United States},
  author={Cramer, Estee Y and Ray, Evan L and Lopez, Velma K and Bracher, Johannes and Brennen, Andrea and Castro Rivadeneira, Alvaro J and Gerding, Aaron and Gneiting, Tilmann and House, Katie H and Huang, Yuxin and others},
  journal={Proceedings of the National Academy of Sciences},
  volume={119},
  number={15},
  pages={e2113561119},
  year={2022},
  publisher={National Academy of Sciences}
}

@article{niu2024multi,
  title={Multi-fidelity residual neural processes for scalable surrogate modeling},
  author={Niu, Ruijia and Wu, Dongxia and Kim, Kai and Ma, Yi-An and Watson-Parris, Duncan and Yu, Rose},
  journal={arXiv preprint arXiv:2402.18846},
  year={2024}
}

@article{ren2025towards,
  title={Towards scientific intelligence: A survey of llm-based scientific agents},
  author={Ren, Shuo and Jian, Pu and Ren, Zhenjiang and Leng, Chunlin and Xie, Can and Zhang, Jiajun},
  journal={arXiv preprint arXiv:2503.24047},
  year={2025}
}

@article{zhang2024sciglm,
  title={Sciglm: Training scientific language models with self-reflective instruction annotation and tuning},
  author={Zhang, Dan and Hu, Ziniu and Zhoubian, Sining and Du, Zhengxiao and Yang, Kaiyu and Wang, Zihan and Yue, Yisong and Dong, Yuxiao and Tang, Jie},
  journal={arXiv preprint arXiv:2401.07950},
  year={2024}
}

@article{yang2024moose,
  title={Moose-chem: Large language models for rediscovering unseen chemistry scientific hypotheses},
  author={Yang, Zonglin and Liu, Wanhao and Gao, Ben and Xie, Tong and Li, Yuqiang and Ouyang, Wanli and Poria, Soujanya and Cambria, Erik and Zhou, Dongzhan},
  journal={arXiv preprint arXiv:2410.07076},
  year={2024}
}

@article{chen2024scienceagentbench,
  title={Scienceagentbench: Toward rigorous assessment of language agents for data-driven scientific discovery},
  author={Chen, Ziru and Chen, Shijie and Ning, Yuting and Zhang, Qianheng and Wang, Boshi and Yu, Botao and Li, Yifei and Liao, Zeyi and Wei, Chen and Lu, Zitong and others},
  journal={arXiv preprint arXiv:2410.05080},
  year={2024}
}

@inproceedings{mialon2023gaia,
  title={Gaia: a benchmark for general ai assistants},
  author={Mialon, Gr{\'e}goire and Fourrier, Cl{\'e}mentine and Wolf, Thomas and LeCun, Yann and Scialom, Thomas},
  booktitle={The Twelfth International Conference on Learning Representations},
  year={2023}
}

@article{farquhar2024detecting,
  title={Detecting hallucinations in large language models using semantic entropy},
  author={Farquhar, Sebastian and Kossen, Jannik and Kuhn, Lorenz and Gal, Yarin},
  journal={Nature},
  volume={630},
  number={8017},
  pages={625--630},
  year={2024},
  publisher={Nature Publishing Group UK London}
}

@article{schick2023toolformer,
  title={Toolformer: Language models can teach themselves to use tools},
  author={Schick, Timo and Dwivedi-Yu, Jane and Dess{\`\i}, Roberto and Raileanu, Roberta and Lomeli, Maria and Hambro, Eric and Zettlemoyer, Luke and Cancedda, Nicola and Scialom, Thomas},
  journal={Advances in Neural Information Processing Systems},
  volume={36},
  pages={68539--68551},
  year={2023}
}

@article{patil2024gorilla,
  title={Gorilla: Large language model connected with massive apis},
  author={Patil, Shishir G and Zhang, Tianjun and Wang, Xin and Gonzalez, Joseph E},
  journal={Advances in Neural Information Processing Systems},
  volume={37},
  pages={126544--126565},
  year={2024}
}

@article{ma2024llm,
  title={Llm and simulation as bilevel optimizers: A new paradigm to advance physical scientific discovery},
  author={Ma, Pingchuan and Wang, Tsun-Hsuan and Guo, Minghao and Sun, Zhiqing and Tenenbaum, Joshua B and Rus, Daniela and Gan, Chuang and Matusik, Wojciech},
  journal={arXiv preprint arXiv:2405.09783},
  year={2024}
}

@article{thulke2024climategpt,
  title={Climategpt: Towards ai synthesizing interdisciplinary research on climate change},
  author={Thulke, David and Gao, Yingbo and Pelser, Petrus and Brune, Rein and Jalota, Rricha and Fok, Floris and Ramos, Michael and Van Wyk, Ian and Nasir, Abdallah and Goldstein, Hayden and others},
  journal={arXiv preprint arXiv:2401.09646},
  year={2024}
}

@article{lewis2020retrieval,
  title={Retrieval-augmented generation for knowledge-intensive nlp tasks},
  author={Lewis, Patrick and Perez, Ethan and Piktus, Aleksandra and Petroni, Fabio and Karpukhin, Vladimir and Goyal, Naman and K{\"u}ttler, Heinrich and Lewis, Mike and Yih, Wen-tau and Rockt{\"a}schel, Tim and others},
  journal={Advances in neural information processing systems},
  volume={33},
  pages={9459--9474},
  year={2020}
}

@article{gao2023retrieval,
  title={Retrieval-augmented generation for large language models: A survey},
  author={Gao, Yunfan and Xiong, Yun and Gao, Xinyu and Jia, Kangxiang and Pan, Jinliu and Bi, Yuxi and Dai, Yixin and Sun, Jiawei and Wang, Haofen and Wang, Haofen},
  journal={arXiv preprint arXiv:2312.10997},
  volume={2},
  number={1},
  year={2023}
}

@article{tian2023just,
  title={Just ask for calibration: Strategies for eliciting calibrated confidence scores from language models fine-tuned with human feedback},
  author={Tian, Katherine and Mitchell, Eric and Zhou, Allan and Sharma, Archit and Rafailov, Rafael and Yao, Huaxiu and Finn, Chelsea and Manning, Christopher D},
  journal={arXiv preprint arXiv:2305.14975},
  year={2023}
}

@article{lin2022teaching,
  title={Teaching models to express their uncertainty in words},
  author={Lin, Stephanie and Hilton, Jacob and Evans, Owain},
  journal={arXiv preprint arXiv:2205.14334},
  year={2022}
}

@article{min2023factscore,
  title={Factscore: Fine-grained atomic evaluation of factual precision in long form text generation},
  author={Min, Sewon and Krishna, Kalpesh and Lyu, Xinxi and Lewis, Mike and Yih, Wen-tau and Koh, Pang Wei and Iyyer, Mohit and Zettlemoyer, Luke and Hajishirzi, Hannaneh},
  journal={arXiv preprint arXiv:2305.14251},
  year={2023}
}

@inproceedings{lopez2018microscopic,
  title={Microscopic traffic simulation using sumo},
  author={Lopez, Pablo Alvarez and Behrisch, Michael and Bieker-Walz, Laura and Erdmann, Jakob and Fl{\"o}tter{\"o}d, Yun-Pang and Hilbrich, Robert and L{\"u}cken, Leonhard and Rummel, Johannes and Wagner, Peter and Wie{\ss}ner, Evamarie},
  booktitle={2018 21st International Conference on Intelligent Transportation Systems (ITSC)},
  pages={2575--2582},
  year={2018},
  organization={IEEE}
}

@inproceedings{yao2023react,
  title={ReAct: Synergizing Reasoning and Acting in Language Models},
  author={Yao, Shunyu and Zhao, Jeffrey and Yu, Dian and Du, Nan and Shafran, Izhak and Narasimhan, Karthik and Cao, Yuan},
  booktitle={International Conference on Learning Representations},
  year={2023},
  url={https://openreview.net/forum?id=WE_vluYUL-X}
}
\bibliographystyle{plain}
\appendix
\section{Appendix}

\subsection{SimulRAG algorithm}
\label{sec:appendix_algorithm}

The SimulRAG algorithm is shown in Algorithm~\ref{alg:simulrag_framework}.
The key functions used in the algorithm are defined as follows:
\begin{itemize}
\item $I(S, q, h)$: Simulator retrieval interface that extracts parameters
from question $q$ using handbook $h$, executes simulator $S$, and converts
outputs to textual context $d$.
\item $\text{LLM}(q)$: Generates a diverse initial answer for question $q$.
\item $\text{Decompose}(a_j)$: Decomposes answer $a_j$ into atomic claims following
factual statement extraction principles.
\item $\text{Merge}(\cdot)$: Merges and deduplicates claim sets from multiple
answers using semantic equivalence detection.
\item $\text{UE}(c_i, \mathcal{A}, \mathcal{C})$: Uncertainty estimation that computes
confidence score for claim $c_i$ using entailment graph centrality.
\item $\text{bound}(c_i, h)$: Boundary assessment function that determines
whether claim $c_i$ is verifiable by simulator using handbook $h$.
\item $\text{VerifyAndUpdate}(c_i, d)$: Verifies claim $c_i$ against simulation
context $d$, retaining it if aligned and updating it if contradicted.
\item $\text{GenerateCoherentAnswer}(q, \mathcal{C}_{\text{final}})$: Synthesizes
final answer from high-confidence claims in response to question $q$.
\end{itemize}

\begin{table}[H]
\centering
\small
\setlength{\tabcolsep}{5pt}
\begin{tabular}{lccccc}
\toprule
\textbf{Domain} & $m$ & $\tau$ & $\kappa$ & \makecell{Tool/SBA/decomp.\\temperature} & \makecell{Initial-answer\\temperature} \\
\midrule
Climate & 5 & 0.4 & 0.4 & 0.1 & 1.0 \\
Epidemiology & 5 & 0.4 & 0.4 & 0.1 & 1.0 \\
Urban & 5 & 0.4 & 0.4 & 0.1 & 1.0 \\
\bottomrule
\end{tabular}
\caption{Hyperparameters for the main non-ablation experiments. Here, $m$ is the number of initial answers, $\tau$ is the claim-selection threshold, and $\kappa$ is the final claim-filtering threshold.}
\label{tab:main_hyperparameters}
\end{table}

\begin{algorithm}[b!]
\caption{SimulRAG framework}
\label{alg:simulrag_framework}
\begin{algorithmic}[1]
\STATE \textbf{Input:} Question $q$, Scientific simulator $S$, Handbook $h$, Thresholds $\tau, \kappa$
\STATE \textbf{Output:} Refined answer $a'$
\STATE $\mathcal{A} \leftarrow \{a_1, \ldots, a_m\}$ where $a_j \leftarrow \text{LLM}(q)$
\FOR{$j = 1$ to $m$}
    \STATE $\{c_{j1}, \ldots, c_{jn_j}\} \leftarrow \text{Decompose}(a_j)$
\ENDFOR
\STATE $\mathcal{C} \leftarrow \text{Merge}(\bigcup_j \{c_{j1}, \ldots, c_{jn_j}\})$
\STATE $\mathcal{C}_{\text{selected}} \leftarrow \emptyset$
\FOR{$c_i \in \mathcal{C}$}
    \STATE $\text{conf}(c_i) \leftarrow \text{UE}(c_i, \mathcal{A}, \mathcal{C})$
    \IF{$\text{conf}(c_i) < \tau$ \AND $\text{bound}(c_i, h) = 1$}
        \STATE $\mathcal{C}_{\text{selected}} \leftarrow \mathcal{C}_{\text{selected}} \cup \{c_i\}$
    \ENDIF
\ENDFOR
\STATE $d \leftarrow I(S, q, h)$
\FOR{$c_i \in \mathcal{C}_{\text{selected}}$}
    \STATE $c_i \leftarrow \text{VerifyAndUpdate}(c_i, d)$
    \STATE $\text{conf}(c_i) \leftarrow 1$
\ENDFOR
\STATE $\mathcal{C}_{\text{final}} \leftarrow \{c_i \in \mathcal{C} \mid \text{conf}(c_i) \geq \kappa\}$
\STATE $a' \leftarrow \text{GenerateCoherentAnswer}(q, \mathcal{C}_{\text{final}})$
\STATE \textbf{return} $a'$
\end{algorithmic}
\end{algorithm}

\subsection{Simulator retrieval interface details}
\label{sec:appendix_retrieval_interface}
The simulator retrieval interface first provides the simulator handbook $h$ together with question $q$ to the tool-calling LLM, which returns simulator arguments in a predefined JSON schema. We discard outputs that fail JSON parsing, use field names inconsistent with the handbook, or contain values outside the valid input ranges, and then re-prompt the model until a valid call is obtained. After validation, the simulator is executed once and the structured output is converted into textual context using the domain template.

\begin{algorithm}[t]
\caption{Simulator retrieval interface}
\label{alg:retrieval_interface}
\begin{algorithmic}[1]
\STATE \textbf{Input:} Question $q$, simulator $S$, handbook $h$
\STATE \textbf{Output:} Textual simulator context $d$
\STATE $\text{valid} \leftarrow 0$
\WHILE{$\text{valid} = 0$}
    \STATE $\hat{p} \leftarrow \text{LLM}_{0.1}(q, h)$ \COMMENT{Generate JSON tool call}
    \IF{$\hat{p}$ is valid JSON, matches handbook field names, and all values are in range}
        \STATE $\text{valid} \leftarrow 1$
    \ENDIF
\ENDWHILE
\STATE $o \leftarrow S(\hat{p})$
\STATE $d \leftarrow \text{TemplateToText}(q, \hat{p}, o)$
\STATE \textbf{return} $d$
\end{algorithmic}
\end{algorithm}

We evaluated the retrieval interface on 200 GPT-4o tool-calling test cases per dataset. Retries were triggered 17 times for climate, 9 times for epidemiology, and 4 times for urban planning, corresponding to 30 retries over 600 calls (5.0\%). Among these retries, 3 were caused by malformed JSON outputs, 23 by variable names that did not strictly follow the handbook specification, and 4 by invalid input values. These results indicate that handbook-aligned field naming is the dominant failure mode, while outright formatting failures are relatively rare.

\subsection{Prompt and template details}
\label{sec:appendix_prompt_template}
Here we provide the prompts and templates used in SimulRAG. Unless otherwise noted, the tool caller, claim decomposer, and scientific boundary analyzer use GPT-4o with a temperature of 0.1. For the multiple initial responses sampled in Algorithm~\ref{alg:simulrag_framework}, we use a temperature of 1.0.

\subsubsection{Parameter extraction prompt}
The tool caller uses the following generic parameter-extraction prompt,
with simulator-specific tool descriptions and parameter schemas filling
the placeholders.

\begin{promptbox}
\begin{adjustwidth}{1em}{0em}
\fontfamily{cmtt}\selectfont\raggedright
You are an expert AI system that translates natural language questions about scientific forecasting into structured parameters for the simulation tool.

**Your Task:**

Given an open-ended question in a scientific domain, extract the precise parameters needed to run the simulator.

**Available Tools:** \{AVAILABLE TOOLS\}

**Available Parameters:** \{AVAILABLE PARAMETERS\}

**Available Target Metrics:** \{AVAILABLE TARGETS\}

**Output Format:** A single JSON object with the specified keys.
\end{adjustwidth}
\end{promptbox}

% \subsubsection{Single Question Answer Prompt}
% This prompt is used to generate one answer a time given the open-ended question. Our framework would use this prompt multiple times to generate the answer set for claim decomposition. The prompt is given as follows:

% \begin{promptbox}
% \begin{adjustwidth}{1em}{0em}
% \fontfamily{cmtt}\selectfont\raggedright % Switch to Computer Modern Typewriter font and left alignment
% Please provide a comprehensive answer to the following climate-related question. 
% Focus on providing factual, well-reasoned information based on [Dataset]'s domain knowledge.
% \end{adjustwidth}
% \end{promptbox}

\subsubsection{Claim decomposition prompt}
After obtaining the answer set, we use the claim decomposition prompt to decompose each answer into atomic claims. It preserves the original meaning while focusing on statements specific to the answer rather than restating the question. We adapt the prompt from Jiang et al.~\citep{jiang2024graph}:

\begin{promptbox}
\begin{adjustwidth}{1em}{0em}
\fontfamily{cmtt}\selectfont\raggedright 
Please deconstruct the following paragraph into the smallest possible standalone self-contained facts without semantic repetition, and return the output as a jsonl, where each line is 
\{claim:[CLAIM]\}.

CRITICAL: Extract ONLY the 8-12 MOST IMPORTANT claims. Be extremely selective. Focus ONLY on:

- Direct answers to the specific question asked

- Specific numerical values, percentages\%, or measurements

- Key causal relationships (A causes B)

- Critical scientific conclusions

STRICTLY AVOID:

- General background information

- Basic definitions (what SO2 is, what SSP scenarios are, etc.)

- Procedural explanations

- Location descriptions

- Any claim that doesn't directly address the question

Each claim must be essential to answering the question. If unsure whether to include a claim, DON'T include it.

The input is: \{original\_text\}

Response:
\end{adjustwidth}
\end{promptbox}

\subsubsection{Claim merging prompt}
Each answer is decomposed into a claim set. We use the claim merging prompt to combine these sets, merge semantically duplicate claims, and retain mutually independent claims. We adapt its design from Jiang et al.~\citep{jiang2024graph}:

\begin{promptbox}
\begin{adjustwidth}{1em}{0em}
\fontfamily{cmtt}\selectfont\raggedright 
You are given two sets of claims. Find which claims in Set B are already covered by claims in Set A.

Set A (Existing claims):
\{existing\_claim\_set\}

Set B (New claims):
\{new\_claim\_set\}

For each claim in Set B, check if it says essentially the same thing as any claim in Set A (i.e. semantic equivalence even if worded differently).
Should be equivalent in meaning, if A said something turns up while B said something turns down, they are not equivalent. If A said something turns up and B said the same thing goes up, they are equivalent.

You must respond with ONLY a valid Json array of pairs. Each pair is [existing\_index, new\_index] where Set A[existing\_index] covers Set B[new\_index].

Examples:

- If Set A[0] covers Set B[1] and Set A[2] covers Set B[0]: [[0, 1], [2, 0]]

- If no claims match: []

- If Set A[1] covers Set B[0]: [[1, 0]]

Response:
\end{adjustwidth}
\end{promptbox}

\subsubsection{Entailment graph construction}
After semantic-equivalent claims are merged, we construct an unweighted
bipartite graph between sampled answers and merged claims. For every
answer--claim pair $(a_j,c_i)$, the LLM entailment judge determines
whether the answer semantically supports the standalone claim using the
same equivalence criteria above and temperature 0.1. We add a binary edge
$(a_j,c_i)$ only for an affirmative judgment; no token-probability threshold
is used. Closeness centrality is then computed on this graph as defined in
Section~\ref{subsec:claim_level_generation}. This binary prompting scheme
keeps the procedure compatible with black-box API models.

\subsubsection{Scientific boundary analysis prompt}
As described in Section~\ref{subsec:claim_level_generation}, the scientific boundary analyzer determines whether a claim can be verified or updated using the tools specified in the simulator handbook. We use the following prompt:

\begin{promptbox}
\begin{adjustwidth}{1em}{0em}
\fontfamily{cmtt}\selectfont\raggedright 
You are an expert in [Dataset] science and computational tools. You will evaluate whether [Dataset]-related claims can be verified using available [Dataset] simulation tools.

**AVAILABLE CLIMATE TOOLS:**
\{tools\_handbook\}

**EVALUATION TASK:**
For each claim provided, you need to determine how well the available tools can help verify or assess the accuracy of that claim.

**SCORING CRITERIA:**

- **0**: The claim cannot be verified or assessed using any of the available tools

- Examples: Claims about non-climate topics, general policy statements, claims requiring data/tools not available, claims that touch on climate aspects but cannot be directly verified with the specific tools provided
  
- **1**: The claim can be directly and comprehensively verified or assessed using the available tools

- Examples: Claims about temperature changes, climate scenarios, aerosol/greenhouse gas impacts, geographic classifications, specific quantitative climate predictions

**RESPONSE FORMAT:**
Respond with ONLY a JSON object containing a single key "tool\_confidence" with a value of 0 or 1.

Example: \{"tool\_confidence": 1\}

**YOUR TASK:**

Question: \{question\}
Claim: \{claim\}

Evaluate how well the available climate simulation tools can verify or assess this specific claim in the context of the given question.

Response:
\end{adjustwidth}
\end{promptbox}

\subsubsection{Verification and updating prompt}

The verification and updating prompt asks the model to verify each selected claim against the textual simulation context and update it when necessary:

\begin{promptbox}
\begin{adjustwidth}{1em}{0em}
\fontfamily{cmtt}\selectfont\raggedright 
You are a fact-checking assistant. You have been given a claim and some quantitative context information. Your task is to analyze the relationship between the claim and the RAG context to see if you should update the claim or not.
You should assume the RAG context is 100\% correct and accurate.

INSTRUCTIONS:

1. First, determine if the RAG context contains information relevant to the claim's topic (set "is\_included")

2. If relevant, check if the claim should be updated for better accuracy (set "should\_update")

3. If updating, modify the claim directly - do not generate new claims or unrelated content. 

4. Only update the related part of the claim, don't add extra information (i.e. if the claim is related to A, and the rag provides information about all of A, B, C, then you should update the parts related to A only, but not B or C)

5. Keep updates as minimal as possible and focused on improving accuracy

6. You may need to do calculations from the RAG context to perform the update, it is required to do and please carefully do the numerical calculations.

7. Skip the update and mark it as not included if the claim is not related to the RAG context.

DECISION CRITERIA:

- "is\_included": true if RAG context discusses the same topic/concept/domain as the claim, false if completely unrelated

- "should\_update": true only if the claim has incorrect/incomplete information that RAG context can improve

Respond in JSON format:
\{\{
    "is\_included": true/false,
    "should\_update": true/false,
    "updated\_claim": "the updated claim text (only if should\_update is true)"
\}\}

YOUR TASK INPUT:
CLAIM TO EVALUATE: \{claim\}

RAG CONTEXT:
\{textual\_context\}

Response:
\end{adjustwidth}
\end{promptbox}

\subsubsection{Final answer prompt}

The final answer prompt asks the model to generate an answer from the selected claims:

\begin{promptbox}
\begin{adjustwidth}{1em}{0em}
\fontfamily{cmtt}\selectfont\raggedright 
You are an expert answering a complex question based on provided factual claims.

QUESTION: \{question\}

AVAILABLE CLAIMS:
\{claims\_text\}

TASK: Generate a comprehensive and accurate answer to the question using only the information provided in the claims above.

REQUIREMENTS:

- Use only the factual information from the provided claims

- Synthesize the claims into a coherent, well-structured answer

- If claims conflict, prioritize the most specific and detailed information

- If the claims don't fully address the question, acknowledge the limitations

- Do not add information beyond what's provided in the claims

Generate a clear, comprehensive answer:
\end{adjustwidth}
\end{promptbox}

\newpage
\subsubsection{Textual context template}
As described in Section~\ref{subsec:simulator_retrieval}, textual context templates convert simulation results into human-readable context by filling placeholder fields with simulator inputs and outputs. A climate template is shown below:

\begin{promptbox}
\begin{adjustwidth}{1em}{0em}
\fontfamily{cmtt}\selectfont\raggedright 
"Query": "If CO2 emissions increase by \{\{delta\_CO2\}\}\% and CH4 emissions increase by \{\{delta\_CH4\}\}\% in \{\{year\}\} under the \{\{setting\}\} scenario, what would be the average temperature for \{\{city\_name\}\}? Also, is \{\{city\_name\}\} located on land or sea?"

"Result": "With a \{\{delta\_CO2\}\}\% increase in CO2 and \{\{delta\_CH4\}\}\% increase in CH4, the average temperature for \{\{city\_name\}\} in \{\{year\}\} under the \{\{setting\}\} scenario would be \{\{greenhouse\_temp\}\}°C. This location is on \{\{land\_sea\_result\}\}."        
\end{adjustwidth}
\end{promptbox}

An epidemiology template is shown below:
\begin{promptbox}
\begin{adjustwidth}{1em}{0em}
\fontfamily{cmtt}\selectfont\raggedright 
"Query": "What is the projected epidemiological landscape for \{\{target\_metric\}\} in \{\{target\_states\}\} for an influenza season initiating around \{\{starting\_date\}\}, assuming a basic reproduction number (R0) of \{\{r0\_value\}\}, a \{\{seasonality\_level\}\} influence, and an initial population immunity level of \{\{prior\_immunity\_level\}\}?"

"Result": "Projected Epidemiological Landscape for \{\{target\_metric\}\}:\{\{simulation\_outlook\}\}"
\end{adjustwidth}
\end{promptbox}

\subsection{Claim-level error analysis}
\label{sec:appendix_error_analysis}
We conducted a granular error analysis over approximately 21{,}138 atomic claims produced in our experiments. Explicit errors were rare, occurring about 165 times in total (approximately 0.78\%). The dominant error category is \textbf{Claim Update} (about 70\%), where the model retrieves the relevant simulation output but misinterprets the numerical significance or context when rewriting the claim. The second category is \textbf{Verification Hallucination} (about 25\%), where the model incorrectly judges a claim as verifiable or unverifiable, corresponding to SBA errors. The remaining errors are \textbf{Parameter Extraction} failures (about 5\%), where the retrieval interface cannot map underspecified questions to precise simulator inputs.

\subsection{Dataset statistics and examples}
\label{sec:appendix_dataset_statistics}
We include examples of questions, ground-truth answers, and the corresponding
ground-truth claim sets from the three benchmark datasets, together with dataset statistics.

\subsubsection{Example climate questions}
Below, we provide example questions from the climate benchmark:

% --- Example 1 ---
\begin{examplebox}
\textbf{Question}: For the city of Chaiwu, what is the total temperature change projected between its historical average in 1994 and a future scenario in 2041 under ssp585 conditions, where CO2 emissions increase by 47.31\% and CH4 emissions increase by 36.46\%? Does this change represent a warming or cooling trend, and would its magnitude be considered modest (less than 1.0\textdegree C) or significant (1.0\textdegree C or greater)?

\vspace{2ex} % Adds a little vertical space for readability

\textbf{Reference Answer}: The total projected temperature change is an increase of 1.08\textdegree C, which indicates a clear warming trend for Chaiwu between the two dates. Based on the provided thresholds, the magnitude of this change is considered significant. This suggests a notable shift in the local climate under the specified emissions scenario.

\vspace{2ex}

\textbf{Reference Key Claims}:
\begin{itemize}[leftmargin=*]
    \item The total projected temperature change is an increase of 1.08\textdegree C.
    \item There is a clear warming trend for Chaiwu between the two dates.
    \item The magnitude of the temperature change is considered significant based on the provided thresholds.
    \item There is a notable shift in the local climate under the specified emissions scenario.
\end{itemize}
\end{examplebox}

% --- Example 2 ---
\begin{examplebox}
\textbf{Question}: For Suresnes in 2050, under the ssp245 scenario with CO2 emissions increasing by 23.83\% and CH4 by 39.2\%, what is the projected temperature? Based on this projection, would the local climate be considered cold (below 5\textdegree C), mild (5\textdegree C to 15\textdegree C), or warm (above 15\textdegree C)? Also, confirm if the location is terrestrial.

\vspace{2ex}

\textbf{Reference Answer}: The projected temperature for Suresnes is 7.9\textdegree C, which is classified as mild according to the given temperature ranges. The location is confirmed to be terrestrial. The projected climate does not fall into the cold or warm categories.

\vspace{2ex}

\textbf{Reference Key Claims}:
\begin{itemize}[leftmargin=*]
    \item The projected temperature for Suresnes is 7.9\textdegree C.
    \item The projected temperature of 7.9\textdegree C for Suresnes is classified as mild.
    \item The projected climate for Suresnes does not fall into the cold category.
    \item The projected climate for Suresnes does not fall into the warm category.
\end{itemize}
\end{examplebox}

% --- Example 3 ---
\begin{examplebox}
\textbf{Question}: For the city of Baldwin in 2072 under the ssp245 scenario, compare two distinct emission modification strategies: one where CO2 emissions increase by 13.4\% and CH4 emissions increase by 6.09\%, and another where SO2 emissions decrease by 13.28\% and Black Carbon emissions are modified by -11.75\% at [(-73.6075, 40.6511)]. Which strategy results in a warmer local climate? What is the temperature difference between the two scenarios? Is this difference considered negligible (less than 0.2\textdegree C), modest (0.2\textdegree C to 1.0\textdegree C), or significant (over 1.0\textdegree C)?

\vspace{2ex}

\textbf{Reference Answer}: The scenario involving an increase in greenhouse gas emissions results in a warmer local climate compared to the aerosol modification scenario. The temperature difference between the two strategies is 0.26\textdegree C, which is considered a modest divergence based on the given thresholds. This indicates that the specified greenhouse gas increases have a more pronounced warming impact than the aerosol changes in this particular forecast.

\vspace{2ex}

\textbf{Reference Key Claims}:
\begin{itemize}[leftmargin=*]
    \item An increase in greenhouse gas emissions results in a warmer local climate compared to aerosol modification.
    \item The temperature difference between the greenhouse gas emissions scenario and the aerosol modification scenario is 0.26\textdegree C.
    \item The 0.26\textdegree C temperature difference is considered a modest divergence based on the given thresholds.
    \item Greenhouse gas increases have a more pronounced warming impact than aerosol changes in this forecast.
\end{itemize}
\end{examplebox}

% --- Example 4 ---
\begin{examplebox}
\textbf{Question}: For Chongzuo, compare the historical average temperature in 1992 with the projection for 2083 under the ssp245 scenario, assuming a 29.37\% increase in CO2 and a 48.75\% increase in CH4 emissions. What is the total temperature change? Based on this change, does this represent a minor (less than 1\textdegree C), significant (1\textdegree C to 2\textdegree C), or severe (over 2\textdegree C) warming trend?

\vspace{2ex}

\textbf{Reference Answer}: The total projected temperature change is an increase of 1.18\textdegree C, which indicates a distinct warming trend for Chongzuo. Based on the provided thresholds, this level of temperature rise is classified as significant. This suggests a considerable shift in the local climate between the historical and future periods under this emissions scenario.

\vspace{2ex}

\textbf{Reference Key Claims}:
\begin{itemize}[leftmargin=*]
    \item The total projected temperature change is an increase of 1.18\textdegree C for Chongzuo.
    \item The 1.18\textdegree C temperature increase indicates a distinct warming trend for Chongzuo.
    \item This level of temperature rise is classified as significant based on the provided thresholds.
    \item The temperature increase suggests a considerable shift in the local climate between historical and future periods under this emissions scenario.
\end{itemize}
\end{examplebox}

% --- Example 5 ---
\begin{examplebox}
\textbf{Question}: For Correia Pinto in the year 2072 under the ssp585 scenario, compare two potential interventions: an aerosol modification with a 6.24\% change in SO2 and -19.46\% change in BC at points [(-50.4, -27.5833)], versus a greenhouse gas intervention with a 44.57\% change in CO2 and 8.5\% change in CH4. Which intervention leads to a warmer local climate? What is the temperature difference between them? Would this difference be considered negligible (less than 0.2\textdegree C), modest (0.2\textdegree C to 1.0\textdegree C), or significant (over 1.0\textdegree C)?

\vspace{2ex}

\textbf{Reference Answer}: The greenhouse gas intervention results in a warmer local climate compared to the aerosol modification. The temperature difference between the two scenarios is 0.61\textdegree C, which is considered a modest impact based on the provided thresholds. This indicates that the choice between these two strategies has a noticeable effect on the projected local temperature.

\vspace{2ex}

\textbf{Reference Key Claims}:
\begin{itemize}[leftmargin=*]
    \item The greenhouse gas intervention results in a warmer local climate compared to the aerosol modification.
    \item The temperature difference between the greenhouse gas intervention and aerosol modification scenarios is 0.61\textdegree C.
    \item The temperature difference of 0.61\textdegree C is considered a modest impact based on the provided thresholds.
    \item The choice between greenhouse gas intervention and aerosol modification strategies has a noticeable effect on the projected local temperature.
\end{itemize}
\end{examplebox}

\subsubsection{Example epidemiology questions}
Below gives example questions generated in our epidemiology benchmark:

% --- Example 6 ---
\begin{examplebox}
\textbf{Question}: For an upcoming influenza season in North Carolina and Massachusetts commencing around October 4th, 2022, what is the comprehensive epidemiological forecast for hospital prevalence, assuming a high R0 of 2.6, moderate seasonality, and a low 10\% prior immunity? Please assess the expected trajectory, including the outbreak's peak severity, and timing.

\vspace{2ex}

\textbf{Reference Answer}: The forecast indicates an extremely severe and rapidly evolving outbreak. The season is projected to begin with an explosive growth phase. The peak magnitude of hospitalizations is expected to be exceptionally high, reaching a median value of around 11,941. This severe peak is anticipated to materialize very quickly, arriving just 5.4 weeks after the season's onset. This trajectory points to a fast-moving, high-burden wave that will develop with remarkable speed.

\vspace{2ex}

\textbf{Reference Key Claims}:
\begin{itemize}[leftmargin=*]
    \item The forecast indicates an extremely severe and rapidly evolving outbreak.
    \item The season is projected to begin with an explosive growth phase.
    \item The peak magnitude of hospitalizations is expected to reach a median value of around 11,941.
    \item The severe peak is anticipated to materialize 5.4 weeks after the season's onset.
    \item The trajectory points to a fast-moving, high-burden wave.
\end{itemize}
\end{examplebox}

% --- Example 7 ---
\begin{examplebox}
\textbf{Question}: For an influenza season in North Carolina, Missouri, Arizona, and Vermont beginning around October 7th, 2022, what is the comprehensive epidemiological forecast for hospital prevalence, assuming a high basic reproduction number (R0) of 2.2, moderate seasonality, and a low 10\% population-wide prior immunity? Please analyze the outbreak's expected trajectory, considering its peak severity, and timing.

\vspace{2ex}

\textbf{Reference Answer}: The forecast points to a severe and rapidly escalating influenza season. The outbreak is projected to begin with an extremely rapid growth phase. The peak magnitude of hospitalizations is expected to be very high, reaching a median of approximately 2234 cases. This significant peak is anticipated to arrive approximately 7.3 weeks after the season's start. Collectively, these indicators suggest a major and sustained wave of influenza.

\vspace{2ex}

\textbf{Reference Key Claims}:
\begin{itemize}[leftmargin=*]
    \item The influenza season is forecasted to be severe and rapidly escalating.
    \item The peak magnitude of hospitalizations is expected to reach a median of approximately 2234 cases.
    \item The peak of hospitalizations is anticipated to arrive approximately 7.3 weeks after the season's start.
    \item Indicators suggest a major and sustained wave of influenza.
\end{itemize}
\end{examplebox}

% --- Example 8 ---
\begin{examplebox}
\textbf{Question}: For an influenza season in Colorado, North Carolina, Vermont, and South Carolina beginning around September 23, 2022, what is the comprehensive epidemiological forecast for hospital prevalence, assuming a high basic reproduction number (R0) of 2.6, no seasonality, and a low population-wide prior immunity of 10\%? Please analyze the projected trajectory, considering the outbreak's peak severity, and timing.

\vspace{2ex}

\textbf{Reference Answer}: The forecast points to an extremely severe and explosive outbreak scenario. The season is projected to begin with an unprecedentedly rapid growth phase. The peak magnitude of hospitalizations is expected to be exceptionally high, reaching a median value of approximately 4,308 concurrent cases. This severe peak is forecast to arrive very quickly, materializing just 4.1 weeks after the season's onset. Taken together, these indicators suggest a rapid, overwhelming wave of infections that escalates to a severe peak in just over a month.

\vspace{2ex}

\textbf{Reference Key Claims}:
\begin{itemize}[leftmargin=*]
    \item The peak magnitude of hospitalizations is expected to reach a median value of approximately 4,308 concurrent cases.
    \item The severe peak of hospitalizations is forecast to materialize 4.1 weeks after the season's onset.
    \item The outbreak is expected to begin with an unprecedentedly rapid growth phase.
    \item The indicators suggest a rapid, overwhelming wave of infections.
\end{itemize}
\end{examplebox}

% --- Example 9 ---
\begin{examplebox}
\textbf{Question}: What is the comprehensive epidemiological forecast for hospital prevalence in Iowa, Florida, and Michigan, for an influenza season starting around September 22nd, 2022? This scenario assumes a high R0 of 2.2, moderate seasonality, and a low 10\% prior immunity in the population. Please consider severity of the outbreak, and its timing.

\vspace{2ex}

\textbf{Reference Answer}: The epidemiological forecast indicates an exceptionally severe and rapidly developing influenza season. The outbreak is expected to begin with an explosive growth phase. This rapid escalation is projected to culminate in a very high peak of hospitalizations, with a median magnitude of approximately 3115 concurrent cases. The peak of this intense wave is forecast to arrive relatively quickly, about 7.1 weeks after the season's onset. This trajectory suggests a severe, front-loaded epidemic wave with a significant and swift impact.

\vspace{2ex}

\textbf{Reference Key Claims}:
\begin{itemize}[leftmargin=*]
    \item The influenza season is expected to be exceptionally severe and rapidly developing.
    \item The outbreak is expected to begin with an explosive growth phase.
    \item The peak of hospitalizations is projected to have a median magnitude of approximately 3115 concurrent cases.
    \item The peak of the influenza wave is forecast to arrive about 7.1 weeks after the season's onset.
    \item The epidemic wave is expected to be severe and front-loaded with a significant and swift impact.
\end{itemize}
\end{examplebox}

% --- Example 10 ---
\begin{examplebox}
\textbf{Question}: For an upcoming influenza season in South Dakota, Massachusetts, Illinois, and Wyoming, what is the comprehensive epidemiological forecast for hospital prevalence given a start date of September 26, 2022, a high R0 of 2.4, strong seasonality, and an initial population immunity level of 20\%? Please assess the expected trajectory, considering the peak's severity, and its timing.

\vspace{2ex}

\textbf{Reference Answer}: The epidemiological forecast for this scenario indicates a severe and sustained influenza wave. The season is expected to begin with a rapid growth phase. The peak magnitude of hospitalizations is projected to be very high, reaching a median value of approximately 999 cases. This severe peak is anticipated to materialize late in the season, occurring about 11 weeks after the start date. Overall, this trajectory points to a challenging season characterized by a rapid initial spread and a high, delayed peak.

\vspace{2ex}

\textbf{Reference Key Claims}:
\begin{itemize}[leftmargin=*]
    \item The epidemiological forecast indicates a severe and sustained influenza wave.
    \item The influenza season is expected to begin with a rapid growth phase.
    \item The peak magnitude of hospitalizations is projected to reach a median value of approximately 999 cases.
    \item The severe peak in hospitalizations is anticipated to occur about 11 weeks after the start date.
    \item The influenza season is characterized by a rapid initial spread and a high, delayed peak.
\end{itemize}
\end{examplebox}

\subsubsection{Example urban planning questions}
Below gives example questions generated in our urban planning benchmark:

% --- Example 1 ---
\begin{examplebox}
\textbf{Question}: To combat downtown gridlock, the city is proposing a two-part strategy: 1) A 'Low-Speed Zone' that reduces speed limits by 27\% to discourage through-traffic, and 2) Upgrading the main corridor's traffic signal to a 'green wave' priority junction. What is the percentage change in average vehicle idling time, and is this reduction considered marginal (under 25\%), moderate (25-75\%), or profound (over 75\%)? What is the percentage change in total CO2 emissions, and is this reduction considered negligible (under 10\%), significant (10-40\%), or transformative (over 40\%)? Based on these projected outcomes, how effectively does this strategy address both traffic flow efficiency and environmental sustainability goals?

\vspace{2ex}

\textbf{Reference Answer}: The combined strategy delivers exceptional improvements in both traffic efficiency and environmental quality. Average vehicle idling time is projected to decrease by 93.1\%, which is a profound reduction indicating significantly smoother traffic flow and reduced congestion. Total CO2 emissions are also expected to decrease by 33.99\%, representing a significant positive impact on air quality. This intervention successfully achieves substantial co-benefits, effectively combating gridlock and advancing environmental sustainability goals by drastically reducing wasted fuel and emissions.

\vspace{2ex}

\textbf{Reference Key Claims}:
\begin{itemize}[leftmargin=*]
    \item Average vehicle idling time is projected to decrease by 93.1\%.
    \item Total CO2 emissions are expected to decrease by 33.99\%.
    \item The strategy delivers improvements in traffic efficiency.
    \item The strategy delivers improvements in environmental quality.
    \item The intervention reduces congestion.
    \item The intervention reduces wasted fuel.
    \item The intervention reduces emissions.
    \item The intervention advances environmental sustainability goals.
\end{itemize}
\end{examplebox}

% --- Example 2 ---
\begin{examplebox}
\textbf{Question}: For our 'Livable Downtown' initiative, a proposed 'road diet' on the main arterial involves reducing the road from 2 lanes to 1 to create a protected bike lane, and simultaneously lowering the speed limit by 26\%. What is the percentage change in average vehicle travel time? Is this change considered minor (under 5\% increase), moderate (5-15\% increase), or significant (over 15\% increase)? What is the percentage change in total CO2 emissions, and is this reduction considered marginal (under 5\%), notable (5-15\%), or substantial (over 15\%)? Considering both impacts, does this 'road diet' represent an acceptable trade-off for urban planners aiming for a 'Livable Downtown'?

\vspace{2ex}

\textbf{Reference Answer}: The proposed 'road diet' results in a significant increase in average vehicle travel time by 15.68\%, indicating a clear impact on traffic flow for motorists. However, the initiative simultaneously delivers a notable environmental benefit, achieving a 14.91\% reduction in total CO2 emissions. This intervention presents a strategic trade-off, prioritizing sustainable transportation modes and air quality improvements over direct vehicular throughput, which is generally acceptable for achieving the broader goals of a 'Livable Downtown' by fostering a more pedestrian and bike-friendly environment.

\vspace{2ex}

\textbf{Reference Key Claims}:
\begin{itemize}[leftmargin=*]
    \item The proposed 'road diet' results in a 15.68\% increase in average vehicle travel time.
    \item The 'road diet' achieves a 14.91\% reduction in total CO2 emissions.
    \item The 'road diet' prioritizes sustainable transportation modes over direct vehicular throughput.
    \item The 'road diet' aims to improve air quality.
    \item The 'road diet' supports the broader goals of a 'Livable Downtown'.
    \item The 'road diet' fosters a more pedestrian and bike-friendly environment.
\end{itemize}
\end{examplebox}

% --- Example 3 ---
\begin{examplebox}
\textbf{Question}: A major downtown concert is expected to increase evening traffic demand by 22\%. To mitigate the anticipated gridlock, the traffic authority is proposing to activate a 'green wave' priority junction along the main arterial leading to the venue. What is the net percentage change in average travel time? What is the net percentage change in vehicle idling, and would this be considered a minor (under 25\%), significant (25-75\%), or transformative (over 75\%) impact? Based on these projected changes, does the 'green wave' strategy effectively counteract the concert's traffic surge and enhance overall urban mobility?

\vspace{2ex}

\textbf{Reference Answer}: The implementation of the 'green wave' priority junction demonstrates exceptional effectiveness in managing the increased traffic demand. Average travel time is projected to decrease by 27.08\%, indicating that the intervention not only offsets the concert's impact but actively improves traffic flow beyond normal conditions. Furthermore, vehicle idling time is expected to decrease by 93.1\%, representing a transformative reduction that will significantly reduce emissions and improve the driver experience. This strategy successfully counters the anticipated gridlock, ensuring a notably more efficient and sustainable urban mobility experience for the evening.

\vspace{2ex}

\textbf{Reference Key Claims}:
\begin{itemize}[leftmargin=*]
    \item Average travel time is projected to decrease by 27.08\% due to the 'green wave' priority junction.
    \item Vehicle idling time is expected to decrease by 93.1\% with the implementation of the 'green wave' priority junction.
    \item The 'green wave' priority junction intervention offsets the concert's impact on traffic.
    \item The 'green wave' priority junction actively improves traffic flow beyond normal conditions.
    \item The reduction in vehicle idling time will significantly reduce emissions.
    \item The 'green wave' priority junction improves the driver experience by reducing idling time.
    \item The strategy counters anticipated gridlock effectively.
    \item The 'green wave' priority junction ensures a more efficient urban mobility experience for the evening.
\end{itemize}
\end{examplebox}

% --- Example 4 ---
\begin{examplebox}
\textbf{Question}: To create a 'Living Street' on the main commercial avenue, the city plans to reduce the speed limit by 32\% to enhance pedestrian safety. To prevent this from causing major gridlock, they will also upgrade the central traffic light to a priority 'green wave' junction. What is the percentage change in average vehicle idling time, and is this reduction considered minor (under 25\%), significant (25-75\%), or drastic (over 75\%)? What is the percentage change in total CO2 emissions, and is this reduction considered minor (under 10\%), moderate (10-30\%), or substantial (over 30\%)? Does this combined intervention effectively achieve its goals for both traffic efficiency and environmental sustainability?

\vspace{2ex}

\textbf{Reference Answer}: The combined intervention demonstrates exceptional success in optimizing urban mobility and environmental quality. Average vehicle idling time sees a drastic reduction of 93.1\%, indicating a highly efficient traffic flow despite the lower speed limit. Furthermore, total CO2 emissions decrease substantially by 35.25\%. These results highlight that the 'green wave' priority junction effectively mitigates congestion from the speed reduction, leading to significant co-benefits for traffic efficiency, pedestrian safety, and air quality.

\vspace{2ex}

\textbf{Reference Key Claims}:
\begin{itemize}[leftmargin=*]
    \item Average vehicle idling time sees a drastic reduction of 93.1\%.
    \item Total CO2 emissions decrease substantially by 35.25\%.
    \item The 'green wave' priority junction effectively mitigates congestion from the speed reduction.
    \item The intervention leads to significant co-benefits for traffic efficiency.
    \item The intervention leads to significant co-benefits for pedestrian safety.
    \item The intervention leads to significant co-benefits for air quality.
\end{itemize}
\end{examplebox}

% --- Example 5 ---
\begin{examplebox}
\textbf{Question}: The city is piloting a 'Smart Corridor' project on its main arterial road. This involves two simultaneous interventions: a dynamic tolling system expected to reduce peak-hour vehicle demand by 31\%, and the replacement of the main traffic light with a 'green wave' priority junction to improve flow. What is the percentage change in average vehicle idling time? How much do total CO2 emissions change, and is this reduction considered minor (less than 25\%), significant (25\% to 50\%), or transformative (over 50\%)? Does this project deliver on its promise to significantly enhance both traffic efficiency and environmental sustainability?

\vspace{2ex}

\textbf{Reference Answer}: The 'Smart Corridor' project demonstrates an exceptional positive impact on both traffic flow and environmental quality. Average vehicle idling time is dramatically reduced by 93.1\%, signifying a near elimination of stop-and-go congestion and substantial operational efficiencies. Furthermore, the initiative achieves a significant environmental improvement with a 49.67\% decrease in total CO2 emissions, falling squarely within the defined 'significant' reduction category. These combined outcomes confirm that the project is highly successful in delivering on its promise of enhancing urban mobility and fostering greater environmental sustainability.

\vspace{2ex}

\textbf{Reference Key Claims}:
\begin{itemize}[leftmargin=*]
    \item The 'Smart Corridor' project reduces average vehicle idling time by 93.1\%.
    \item The 'Smart Corridor' project decreases total CO2 emissions by 49.67\%.
    \item The reduction in CO2 emissions falls within the 'significant' reduction category.
    \item The 'Smart Corridor' project nearly eliminates stop-and-go congestion.
    \item The 'Smart Corridor' project enhances urban mobility.
    \item The 'Smart Corridor' project fosters greater environmental sustainability.
\end{itemize}
\end{examplebox}

\subsubsection{Dataset statistics}
Table~\ref{tab:benchmark_stats} summarizes the statistics of the three benchmark datasets:

\begin{table}[h!]
\centering
% Add \small to reduce the font size of the table content
\small
% Use a booktabs-style tabular environment
% @{} removes extra whitespace on the left and right sides of the table
% lcccccc... defines the column alignment (left, center, center...)
\resizebox{0.98\textwidth}{!}{%
\begin{tabular}{@{}lcccccc@{}}
\toprule
% Use the \makecell command to break long headers into multiple lines and bold them
\textbf{Benchmark} & \makecell[c]{\# Questions} & \makecell[c]{Avg. Answer \\ Length (words)} & \makecell[c]{Avg. \# Ref. \\ Claims} & \makecell[c]{Avg. \# Quant. \\ Ref. Claims} & \makecell[c]{Avg. \# Qual. \\ Ref. Claims} & \makecell[c]{Avg. Tool \\ Param. Count} \\
\midrule
Climate      & 1000 & 55.8 & 3.6 & 1.1 & 2.5 & 6.3 \\
Epidemiology & 1000 & 82.3 & 5.3 & 3.2 & 2.1 & 6.0 \\
Urban & 200 & 116.8 & 7.8 & 2.0 & 5.8 & 6.2 \\
\bottomrule
\end{tabular}
}
\caption{Statistical overview of the generated benchmarks. The table compares answer length, the average number of reference claims (total, quantitative, and qualitative), and the average number of simulator parameters used.}
\label{tab:benchmark_stats}
\end{table}

\subsection{Additional experimental results}
\label{sec:appendix_experimental_results}

\subsubsection{Tool-first ReAct comparison}
\label{subsec:appendix_react}
We adapt ReAct~\citep{yao2023react} to invoke the same simulator
retrieval interface before answer generation. Short ReAct produces a
single-pass tool-calling answer, whereas Long ReAct is prompted to produce
more claims. Table~\ref{tab:react_comparison} reports this dedicated
comparison. Because it uses a separate matched generation protocol and the
updated Claude model, its absolute values should be interpreted within the
table rather than compared directly with Figure~\ref{fig:simulrag_results}.

Short ReAct often attains high factuality by producing few claims, while
Long ReAct increases informativeness but sharply reduces factuality.
SimulRAG maintains high factuality while providing substantially broader
claim coverage across domains and models.

\begin{table}[H]
\centering
\small
\setlength{\tabcolsep}{5pt}
\begin{tabular}{llccc}
\toprule
\textbf{Domain} & \textbf{Model} & \textbf{Short ReAct} & \textbf{Long ReAct} & \textbf{SimulRAG} \\
\midrule
Climate & GPT-4o & 97.14\% / 4.76 & 39.48\% / 9.04 & 85.38\% / 9.55 \\
Climate & Claude-Haiku-4.5 & 85.43\% / 6.80 & 52.55\% / 11.20 & 87.66\% / 10.82 \\
Epidemiology & GPT-4o & 76.36\% / 6.46 & 55.88\% / 14.72 & 90.24\% / 18.68 \\
Epidemiology & Claude-Haiku-4.5 & 71.97\% / 7.24 & 47.71\% / 13.94 & 91.11\% / 18.48 \\
Urban & GPT-4o & 89.16\% / 6.58 & 75.71\% / 15.38 & 90.97\% / 14.30 \\
Urban & Claude-Haiku-4.5 & 90.18\% / 9.18 & 62.56\% / 18.08 & 94.37\% / 18.78 \\
\bottomrule
\end{tabular}
\caption{Tool-first ReAct baselines versus SimulRAG. Each entry reports factuality / informativeness. Short ReAct favors concise, high-factuality answers, while Long ReAct increases coverage at a substantial factuality cost. SimulRAG provides the strongest overall balance.}
\label{tab:react_comparison}
\end{table}

\subsubsection{Baseline comparison: precision and recall}
\label{subsec:appendix_precision_recall}
Table~\ref{tab:recall_precision_benchmark_stats} reports the precision and recall values
corresponding to the F1 scores in Table~\ref{tab:ue_sba_results_final}.

\begin{table*}[h!]
\centering
\begin{small}
\setlength{\tabcolsep}{4pt}
\begin{tabular}{@{}lll||ccc||ccc@{}}
\toprule
% --- Header Row 1: Main Columns ---
\multirow{2}{*}{Benchmark} & \multirow{2}{*}{Metric} & \multirow{2}{*}{Method} & \multicolumn{3}{c||}{GPT-4o} & \multicolumn{3}{c}{Claude 3.5} \\
\cmidrule(lr){4-6} \cmidrule(l){7-9}
% --- Header Row 2: Thresholds ---
& & & 15\% & 25\% & 45\% & 15\% & 25\% & 45\% \\
\midrule
% --------------------------------------------------------------------------
% --- Climate Benchmark ---
% --------------------------------------------------------------------------
\multirow{12}{*}{Climate} & \multirow{4}{*}{Precision} & Random & 0.6670 & 0.6777 & 0.7043 & 0.7364 & 0.7402 & 0.7496 \\
& & Verbalized & 0.6386 & 0.6569 & 0.6910 & 0.6871 & 0.6958 & 0.7220 \\
& & Uncertainty & 0.6901 & 0.7113 & 0.7433 & 0.7433 & 0.7544 & 0.7926 \\
& & UE+SBA (Ours) & \bf{0.7025} & \bf{0.7323} & \bf{0.7719} & \bf{0.7513} & \bf{0.7746} & \bf{0.8238} \\
\cmidrule{2-9}
& \multirow{4}{*}{Recall} & Random & 0.6700 & 0.6811 & 0.7036 & 0.7283 & 0.7437 & 0.7525 \\
& & Verbalized & 0.6558 & 0.6261 & \bf{0.8046} & 0.6656 & 0.7406 & \bf{0.9075} \\
& & Uncertainty & 0.6881 & 0.7109 & 0.7448 & 0.7372 & 0.7505 & 0.7926 \\
& & UE+SBA (Ours) & \bf{0.7014} & \bf{0.7297} & 0.7731 & \bf{0.7510} & \bf{0.7790} & 0.8246 \\
\cmidrule{2-9}
& \multirow{4}{*}{F1 Score} & Random & 0.6685 & 0.6794 & 0.7039 & 0.7323 & 0.7419 & 0.7511 \\
& & Verbalized & 0.6519 & 0.6470 & 0.7393 & 0.6766 & 0.7251 & 0.7230 \\
& & Uncertainty & 0.6894 & 0.7113 & 0.7440 & 0.7384 & 0.7509 & 0.7930 \\
& & UE+SBA (Ours) & \bf{0.7020} & \bf{0.7310} & \bf{0.7725} & \bf{0.7511} & \bf{0.7768} & \bf{0.8242} \\
\midrule
% --------------------------------------------------------------------------
% --- Epidemiology Benchmark ---
% --------------------------------------------------------------------------
\multirow{12}{*}{Epidemiology} & \multirow{4}{*}{Precision} & Random & 0.5910 & 0.6128 & 0.6580 & 0.6945 & 0.7057 & 0.7353 \\
& & Verbalized & 0.5610 & 0.5891 & 0.7021 & 0.6806 & 0.7018 & 0.7565 \\
& & Uncertainty & 0.6173 & 0.6470 & 0.7139 & 0.7138 & 0.7387 & 0.7863 \\
& & UE+SBA (Ours) & \bf{0.6416} & \bf{0.6952} & \bf{0.8131} & \bf{0.7233} & \bf{0.7579} & \bf{0.8200} \\
\cmidrule{2-9}
& \multirow{4}{*}{Recall} & Random & 0.5886 & 0.6185 & 0.6582 & 0.6977 & 0.7026 & 0.7395 \\
& & Verbalized & 0.5735 & 0.6589 & 0.6834 & 0.6866 & 0.7937 & 0.7065 \\
& & Uncertainty & 0.6164 & 0.6484 & 0.7126 & 0.7027 & 0.7393 & 0.7854 \\
& & UE+SBA (Ours) & \bf{0.6445} & \bf{0.6962} & \bf{0.8178} & \bf{0.7229} & \bf{0.7608} & \bf{0.8214} \\
\cmidrule{2-9}
& \multirow{4}{*}{F1 Score} & Random & 0.5898 & 0.6157 & 0.6581 & 0.6961 & 0.7042 & 0.7374 \\
& & Verbalized & 0.5654 & 0.6270 & 0.6953 & 0.6804 & 0.7421 & 0.7310 \\
& & Uncertainty & 0.6103 & 0.6440 & 0.7043 & 0.7060 & 0.7319 & 0.7848 \\
& & UE+SBA (Ours) & \bf{0.6431} & \bf{0.6957} & \bf{0.8155} & \bf{0.7231} & \bf{0.7594} & \bf{0.8207} \\
\bottomrule
\midrule
% --------------------------------------------------------------------------
% --- Urban Benchmark (New) ---
% --------------------------------------------------------------------------
\multirow{12}{*}{Urban} & \multirow{4}{*}{Precision} & Random & 0.7913 & 0.7969 & 0.8085 & 0.6776 & 0.6996 & 0.7493 \\
& & Verbalized & 0.7777 & 0.7859 & 0.8022 & 0.6863 & 0.7105 & 0.7545 \\
& & Uncertainty & 0.7949 & 0.8013 & 0.8210 & 0.6857 & 0.7063 & 0.7580 \\
& & UE+SBA (Ours) & \bf{0.8128} & \bf{0.8231} & \bf{0.8344} & \bf{0.7163} & \bf{0.7624} & \bf{0.8218} \\
\cmidrule{2-9}
& \multirow{4}{*}{Recall} & Random & 0.7925 & 0.7990 & 0.8124 & 0.6826 & 0.7065 & 0.7557 \\
& & Verbalized & 0.7778 & 0.7859 & 0.8022 & 0.6864 & 0.7105 & 0.7546 \\
& & Uncertainty & 0.7949 & 0.7973 & 0.8092 & 0.6821 & 0.7102 & 0.7497 \\
& & UE+SBA (Ours) & \bf{0.8028} & \bf{0.8216} & \bf{0.8236} & \bf{0.7182} & \bf{0.7602} & \bf{0.8225} \\
\cmidrule{2-9}
& \multirow{4}{*}{F1 Score} & Random & 0.7919 & 0.7980 & 0.8105 & 0.6801 & 0.7030 & 0.7525 \\
& & Verbalized & 0.7777 & 0.7859 & 0.8022 & 0.6863 & 0.7105 & 0.7545 \\
& & Uncertainty & 0.7949 & 0.7993 & 0.8151 & 0.6839 & 0.7082 & 0.7538 \\
& & UE+SBA (Ours) & \bf{0.8078} & \bf{0.8224} & \bf{0.8290} & \bf{0.7173} & \bf{0.7613} & \bf{0.8222} \\
\bottomrule
\end{tabular}
\end{small}
\caption{Additional claim-level generation results. The table reports precision, recall, and F1 score at verification budgets of 15\%, 25\%, and 45\%; F1 scores match those in Table~\ref{tab:ue_sba_results_final}.}
\label{tab:recall_precision_benchmark_stats}
\end{table*}

\subsubsection{Detailed performance comparison with confidence intervals}
\label{subsec:appendix_confidence_intervals}
Table~\ref{tab:confidence_intervals_full} reports F1 score, AUPR, and AUROC with 95\% confidence intervals for the Climate and Epidemiology benchmarks.

\begin{table*}[h!]
\centering
\small
\setlength{\tabcolsep}{3pt}
\resizebox{0.98\textwidth}{!}{%
\begin{tabular}{@{}lll||ccc||ccc@{}}
\toprule
\multirow{2}{*}{\textbf{Benchmark}} & \multirow{2}{*}{\textbf{Metric}} & \multirow{2}{*}{\textbf{Method}} & \multicolumn{3}{c||}{\textbf{GPT-4o}} & \multicolumn{3}{c}{\textbf{Claude 3.5}} \\
\cmidrule(lr){4-6} \cmidrule(l){7-9}
& & & 15\% & 25\% & 45\% & 15\% & 25\% & 45\% \\
\midrule
% Climate
\multirow{9}{*}{Climate} & \multirow{3}{*}{F1 Score} & Random & 0.669 $\pm$ 0.05 & 0.679 $\pm$ 0.05 & 0.704 $\pm$ 0.04 & 0.732 $\pm$ 0.03 & 0.742 $\pm$ 0.05 & 0.751 $\pm$ 0.05 \\
& & Uncertainty & 0.689 $\pm$ 0.02 & 0.712 $\pm$ 0.02 & 0.744 $\pm$ 0.03 & 0.740 $\pm$ 0.04 & 0.753 $\pm$ 0.03 & 0.793 $\pm$ 0.05 \\
& & UE+SBA & \bf{0.702 $\pm$ 0.03} & \bf{0.731 $\pm$ 0.00} & \bf{0.773 $\pm$ 0.02} & \bf{0.751 $\pm$ 0.05} & \bf{0.777 $\pm$ 0.05} & \bf{0.824 $\pm$ 0.04} \\
\cmidrule{2-9}
& \multirow{3}{*}{AUPR} & Random & 0.714 $\pm$ 0.02 & 0.738 $\pm$ 0.02 & 0.759 $\pm$ 0.03 & 0.788 $\pm$ 0.03 & 0.793 $\pm$ 0.02 & 0.805 $\pm$ 0.04 \\
& & Uncertainty & 0.740 $\pm$ 0.03 & 0.754 $\pm$ 0.01 & 0.771 $\pm$ 0.05 & 0.778 $\pm$ 0.06 & 0.784 $\pm$ 0.06 & 0.809 $\pm$ 0.04 \\
& & UE+SBA & \bf{0.746 $\pm$ 0.02} & \bf{0.763 $\pm$ 0.03} & \bf{0.787 $\pm$ 0.02} & \bf{0.788 $\pm$ 0.04} & \bf{0.804 $\pm$ 0.02} & \bf{0.830 $\pm$ 0.04} \\
\cmidrule{2-9}
& \multirow{3}{*}{AUROC} & Random & 0.601 $\pm$ 0.06 & 0.627 $\pm$ 0.06 & 0.662 $\pm$ 0.05 & 0.667 $\pm$ 0.02 & 0.679 $\pm$ 0.02 & 0.703 $\pm$ 0.04 \\
& & Uncertainty & 0.642 $\pm$ 0.02 & 0.667 $\pm$ 0.03 & 0.704 $\pm$ 0.02 & 0.683 $\pm$ 0.05 & 0.703 $\pm$ 0.05 & 0.745 $\pm$ 0.03 \\
& & UE+SBA & \bf{0.653 $\pm$ 0.04} & \bf{0.685 $\pm$ 0.01} & \bf{0.725 $\pm$ 0.02} & \bf{0.700 $\pm$ 0.03} & \bf{0.729 $\pm$ 0.02} & \bf{0.771 $\pm$ 0.03} \\
\midrule
% Epidemiology
\multirow{9}{*}{Epidemiology} & \multirow{3}{*}{F1 Score} & Random & 0.590 $\pm$ 0.06 & 0.616 $\pm$ 0.07 & 0.658 $\pm$ 0.06 & 0.696 $\pm$ 0.04 & 0.704 $\pm$ 0.04 & 0.737 $\pm$ 0.03 \\
& & Uncertainty & 0.617 $\pm$ 0.09 & 0.647 $\pm$ 0.04 & 0.714 $\pm$ 0.04 & 0.708 $\pm$ 0.07 & 0.737 $\pm$ 0.05 & 0.785 $\pm$ 0.03 \\
& & UE+SBA & \bf{0.643 $\pm$ 0.07} & \bf{0.696 $\pm$ 0.04} & \bf{0.816 $\pm$ 0.04} & \bf{0.723 $\pm$ 0.06} & \bf{0.759 $\pm$ 0.05} & \bf{0.821 $\pm$ 0.04} \\
\cmidrule{2-9}
& \multirow{3}{*}{AUPR} & Random & 0.673 $\pm$ 0.03 & 0.705 $\pm$ 0.03 & 0.756 $\pm$ 0.06 & 0.743 $\pm$ 0.04 & 0.769 $\pm$ 0.04 & 0.810 $\pm$ 0.04 \\
& & Uncertainty & 0.695 $\pm$ 0.06 & 0.726 $\pm$ 0.05 & 0.787 $\pm$ 0.06 & 0.755 $\pm$ 0.08 & 0.784 $\pm$ 0.03 & 0.825 $\pm$ 0.02 \\
& & UE+SBA & \bf{0.724 $\pm$ 0.06} & \bf{0.775 $\pm$ 0.07} & \bf{0.842 $\pm$ 0.05} & \bf{0.774 $\pm$ 0.06} & \bf{0.811 $\pm$ 0.03} & \bf{0.859 $\pm$ 0.02} \\
\cmidrule{2-9}
& \multirow{3}{*}{AUROC} & Random & 0.645 $\pm$ 0.02 & 0.675 $\pm$ 0.02 & 0.730 $\pm$ 0.05 & 0.572 $\pm$ 0.04 & 0.605 $\pm$ 0.04 & 0.672 $\pm$ 0.04 \\
& & Uncertainty & 0.691 $\pm$ 0.04 & 0.735 $\pm$ 0.03 & 0.808 $\pm$ 0.04 & 0.609 $\pm$ 0.10 & 0.657 $\pm$ 0.06 & 0.738 $\pm$ 0.02 \\
& & UE+SBA & \bf{0.731 $\pm$ 0.04} & \bf{0.791 $\pm$ 0.05} & \bf{0.863 $\pm$ 0.03} & \bf{0.634 $\pm$ 0.08} & \bf{0.697 $\pm$ 0.05} & \bf{0.792 $\pm$ 0.05} \\
\bottomrule
\end{tabular}
}
\caption{Detailed performance comparison (F1, AUPR, AUROC) with 95\% confidence intervals on the Climate and Epidemiology benchmarks. The Verbalized baseline is omitted for brevity but follows similar trends.}
\label{tab:confidence_intervals_full}
\end{table*}

\subsubsection{Precision-recall curves among UE methods}
The following precision-recall curves compare UE+SBA with baseline uncertainty approaches on the Climate, Epidemiology, and Urban datasets. At a 45\% verification budget, UE+SBA uses fewer than half as many RAG updates as all-RAG while achieving comparable performance.

\begin{figure*}[bp]
\centering
\includegraphics[width=0.8\textwidth]{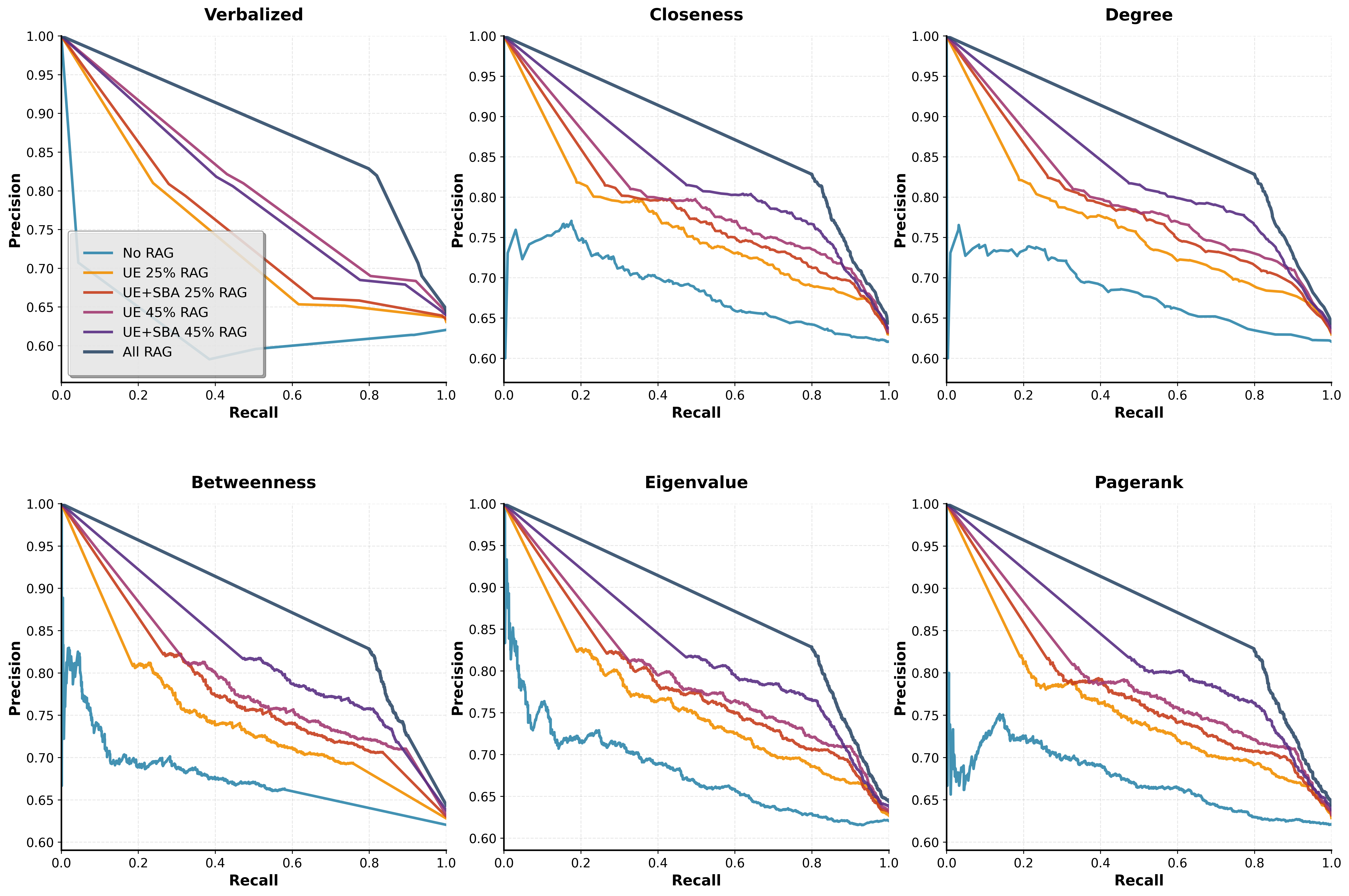}
\caption{Precision-recall curves for Uncertainty and UE+SBA using six uncertainty estimators on the Climate benchmark.}
\label{fig:climate_pr_curves}
\end{figure*}

\begin{figure*}[t]
\centering
\includegraphics[width=0.8\textwidth]{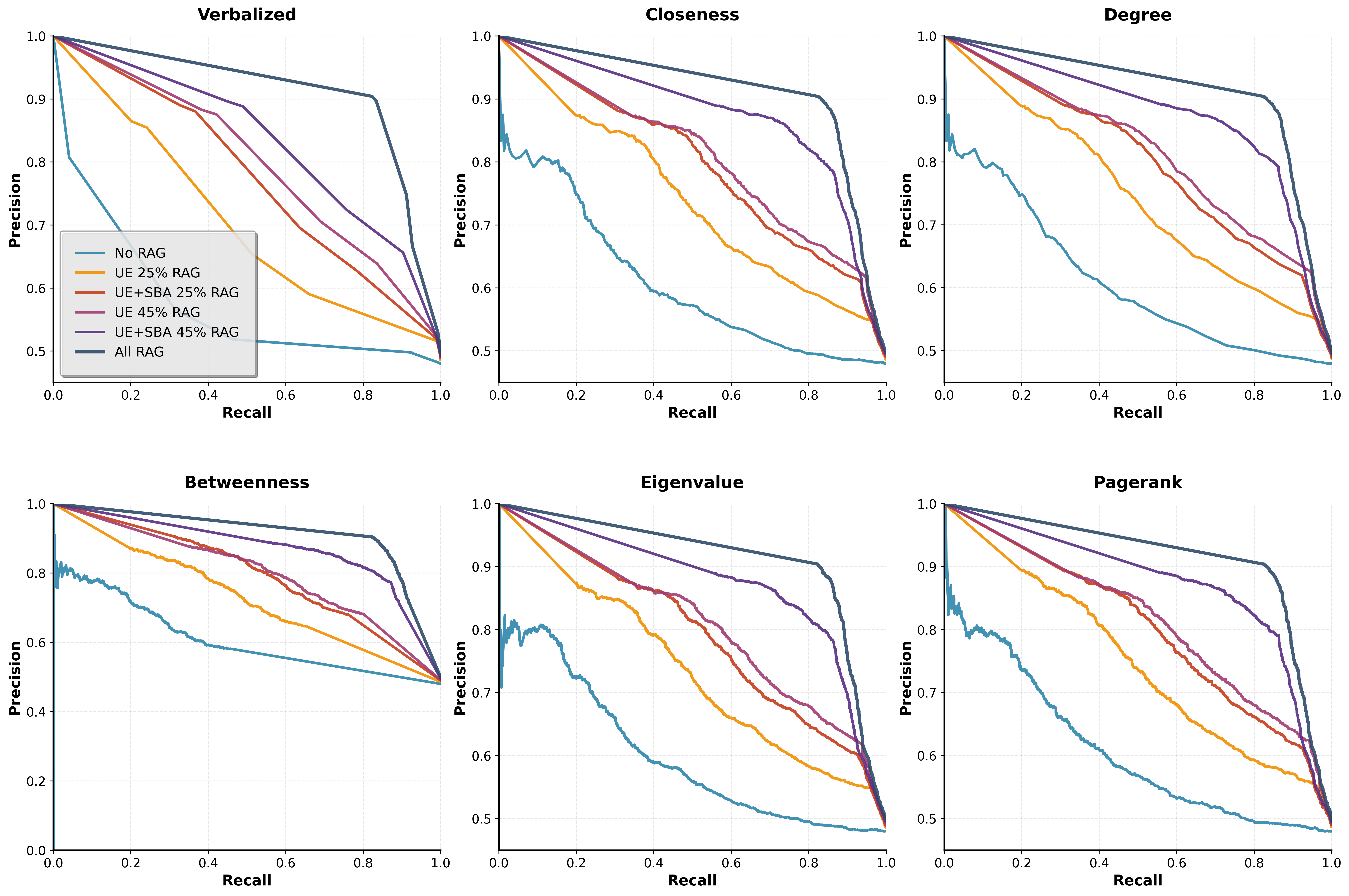}
\caption{Precision-recall curves for Uncertainty and UE+SBA using six uncertainty estimators on the Epidemiology benchmark.}
\label{fig:epidemiology_pr_curves}
\end{figure*}

\begin{figure*}[t]
\centering
\includegraphics[width=0.8\textwidth]{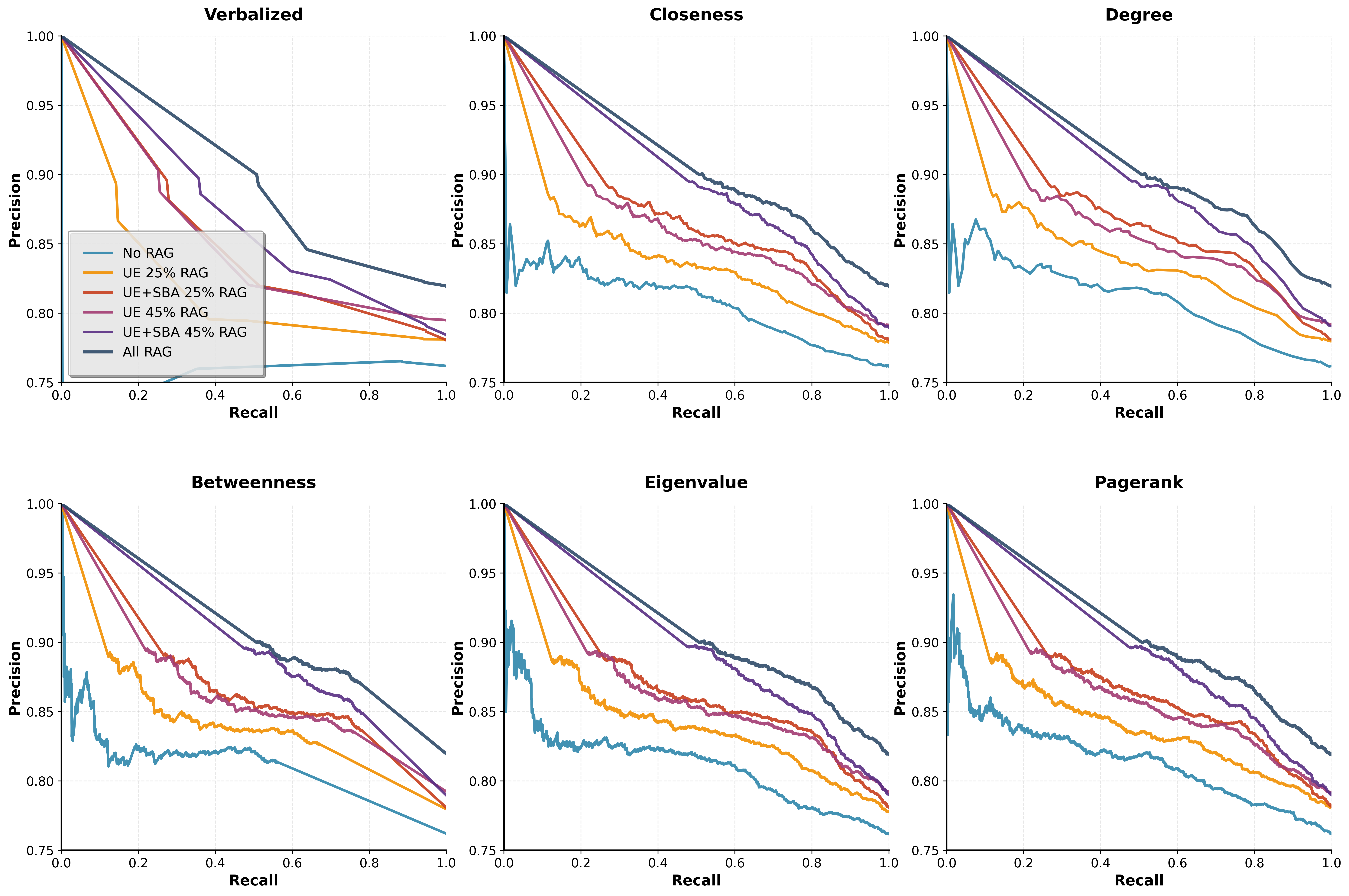}
\caption{Precision-recall curves for Uncertainty and UE+SBA using six uncertainty estimators on the Urban benchmark.}
\label{fig:urban_pr_curves}
\end{figure*}


% --- New Table 2 Added Below ---

% \begin{table*}[t]
% \caption{Comparison with non-RAG and full-RAG methods. This table evaluates performance against baselines without any Retrieval-Augmented Generation (Without RAG) and with RAG applied to all inputs (All RAG).}
% \label{tab:rag_comparison}
% \centering
% \resizebox{\textwidth}{!}{%
% \begin{tabular}{@{}llcccccccccc@{}}
% \toprule
% \multirow{2}{*}{\textbf{Benchmark}} & \multirow{2}{*}{\textbf{Model}} & \multicolumn{2}{c}{\textbf{Without RAG}} & \multicolumn{2}{c}{\textbf{Ours (15\%)}} & \multicolumn{2}{c}{\textbf{Ours (25\%)}} & \multicolumn{2}{c}{\textbf{Ours (45\%)}} & \multicolumn{2}{c}{\textbf{All RAG}} \\
% \cmidrule(lr){3-4} \cmidrule(lr){5-6} \cmidrule(lr){7-8} \cmidrule(lr){9-10} \cmidrule(lr){11-12}
%  &  & AUROC & AUPR & AUROC & AUPR & AUROC & AUPR & AUROC & AUPR & AUROC & AUPR \\ \midrule
% % --- Climate Data ---
% \multirow{2}{*}{Climate} & GPT-4o & 0.5822 & 0.6890 & 0.6593 & 0.8226 & 0.6804 & 0.8356 & 0.7039 & 0.8493 & 0.7206 & 0.8599 \\
%  & Claude 3.5 & 0.0000 & 0.0000 & 0.0000 & 0.0000 & 0.0000 & 0.0000 & 0.0000 & 0.0000 & 0.0000 & 0.0000 \\ \midrule
% % --- Epidemiology Data ---
% \multirow{2}{*}{Epidemiology} & GPT-4o & 0.5399 & 0.6475 & 0.6537 & 0.7727 & 0.6867 & 0.7991 & 0.7227 & 0.8312 & 0.7358 & 0.8435 \\
%  & Claude 3.5 & 0.0000 & 0.0000 & 0.0000 & 0.0000 & 0.0000 & 0.0000 & 0.0000 & 0.0000 & 0.0000 & 0.0000 \\ \bottomrule
% \end{tabular}%
% } % end of resizebox
% \end{table*}

\end{document}